\documentclass{article}

    \usepackage[preprint]{neurips_2020}

\usepackage{microtype}
\usepackage{graphicx}
\usepackage{subfigure}
\usepackage{booktabs} % for professional tables
\usepackage{multirow}
\usepackage{hyperref}
\usepackage{bbm}

\usepackage{amsmath}
\usepackage{amssymb}
\usepackage{mathtools}
\usepackage{amsthm}

\usepackage[capitalize,noabbrev]{cleveref}

\theoremstyle{plain}
\newtheorem{theorem}{Theorem}[section]
\newtheorem{proposition}[theorem]{Proposition}
\newtheorem{lemma}[theorem]{Lemma}

\theoremstyle{definition}

\theoremstyle{remark}

\usepackage[utf8]{inputenc} % allow utf-8 input
\usepackage[T1]{fontenc}    % use 8-bit T1 fonts
\usepackage{hyperref}       % hyperlinks
\usepackage{url}            % simple URL typesetting
\usepackage{booktabs}       % professional-quality tables
\usepackage{amsfonts}       % blackboard math symbols
\usepackage{nicefrac}       % compact symbols for 1/2, etc.
\usepackage{microtype}      % microtypography
\usepackage{graphicx}
 \usepackage{indentfirst} 
 \usepackage{algorithm}
 \usepackage{algorithmic}
\title{Reconstruction Task Finds Universal Winning Tickets}

\author{%
  Ruichen Li\\
  Peking University\\
  \texttt{xk-lrc@pku.edu.cn}\\
  \And
  Binghui Li\\
  Peking University\\
  \texttt{libinghui@pku.edu.cn}\\
  \AND
  Qi Qian\\
  Alibaba Group\\
  \texttt{qi.qian@alibaba-inc.com}\\
  \And
  Liwei Wang\\
  Peking University\\
  \texttt{wanglw@cis.pku.edu.cn}
}
\begin{document}



\maketitle

\begin{abstract}

Pruning well-trained neural networks is effective to achieve a promising accuracy-efficiency trade-off in computer vision regimes. However, most of existing pruning algorithms only focus on the classification task defined on the source domain. Different from the strong transferability of the original model, a pruned network is hard to transfer to complicated downstream tasks such as object detection \cite{girish2021lottery}. In this paper, we show that the image-level pretrain task is not capable of pruning models for diverse downstream tasks. To mitigate this problem, we introduce image reconstruction, a pixel-level task, into the traditional pruning framework. Concretely, an autoencoder is trained based on the original model, and then the pruning process is optimized with both autoencoder and classification losses. The empirical study on benchmark downstream tasks shows that the proposed method can outperform state-of-the-art results explicitly.
%Our experiment results show that our method has a better performance than previous results.

\end{abstract}

\section{Introduction}
\label{Introduction}

Fine-tuning a pre-trained model, which can leverage the knowledge from a large-scale pre-training dataset, becomes prevalent for downstream tasks. This strategy avoids overfitting on small datasets leading to better performance on target tasks. Benefits from the pretrain-finetune strategy, scaling up model capacity is a trend in recent research \cite{pmlr-v139-touvron21a,dosovitskiy2020image,Liu2021swin}. However, large-scale models consume a lot of computational and memory resources, limiting their applications on edge devices. Many efforts are devoted to reducing the computational requirements of neural networks \cite{hubara2017quantized,hinton2015distilling,tai2016convolutional}. Among them, pruning \cite{han2015learning} aims to remove unimportant parameters from the original model and can reduce the size of the model significantly. Most pruning methods rely on a well-trained network and can achieve extraordinary compression rates with negligible accuracy drop on the same task. \cite{han2015learning,yang2019proxsgd,sanh2020movement}

Although pruning methods demonstrate an excellent accuracy vs. sparsity trade-off, only a few works evaluate the pruned model's transferability, i.e., the performance on different downstream tasks. Given multiple downstream tasks, a pruning algorithm can be applied to the individual task. However, the cost that linearly depends on the number of tasks will become intractable. To mitigate this problem, we try to find a pruned model, called universal winning tickets, that can transfer to diverse downstream tasks. 

The lottery tickets hypothesis, proposed by \cite{frankle2018lottery}, claims that each over-parameterized neural network has a sparse subnetwork called a winning ticket, which can achieve the same performance as the entire network. The transferability of winning tickets has been investigated in \cite{morcos2019one}. They show that an ImageNet ticket can transfer to different downstream classification tasks. In \cite{chen2021lottery}, the authors suggest that a pretraining procedure can be regarded as a special initialized method. This kind of initialization is directly amenable to sparsification. Based on this insight, the authors use task agnostic pretraining to help find the universal winning tickets. Their results show that a universal winning ticket exists across different classification downstream tasks. 

However, in some more complicated downstream tasks, such as object detection, tickets found by \cite{chen2021lottery} can result in a degenerated performance. In \cite{chen2021lottery}, the authors reveal that tickets found by the target object detection task surpass tickets found by image classification with a non-negligible margin. In \cite{girish2021lottery}, the authors check tickets found by supervised learning on an object detection dataset \cite{lin2014microsoft}. The result confirms that ImageNet tickets only transfer to a limited extent to downstream tasks, such as object detection or instance segmentation. These observations illustrate that the pruning method's transferability is highly related to the task type. 

In this paper, we aim to find the universal tickets for diverse downstream tasks. Most of the existing methods rely on an image-level task to prune pre-trained models. Although the pruning pipeline has shown an extraordinary performance on downstream tasks \cite{he2020momentum}, it does not treat details and global features in the same status. The implicit tendency of image-level loss causes the neural network to forget pixel-level information during the pruning process. After pruning, pixel-level information becomes untraceable while it is essential for complicated downstream tasks, e.g., detection,  segmentation, etc. The intuition is theoretically analyzed in Section \ref{sec:Theorem}. Therefore, a pixel-level task is necessary for pruning pre-trained models to preserve sufficient information and can help the model transfer to generic downstream tasks.

Unlike image-level tasks, designing appropriate pixel-level tasks is still challenging. Inspired by the recent progress in self-supervised learning \cite{he2021masked}, we introduce the image reconstruction task to find the universal tickets, and a two-stage training paradigm is proposed to obtain the desired ticket. First, an autoencoder structure is introduced for the existing model. The encoder structure inherits the original model structure and weights. Unlike an end-to-end unsupervised pre-training in \cite{he2021masked},  which requires an extremely large model and high mask rate to avoid cheating model, a much smaller decoder is trained in our method for a specific encoder. We use the feature map hint method to accelerate the convergence of decoder. After the first stage of training for the decoder, we have a classification task with a classification head for the second stage of training. Concretely, we freeze the decoder and apply a modified LTH algorithm to get a universal ticket. Finally, the performance is evaluated by transferring the obtained tickets to different downstream tasks.

Our contributions can be summarized as follows.
\begin{itemize}
    \item We propose a new framework for pruning pre-trained neural networks. Different from directly pruning on the classification tasks, we first train a decoder for the pruned network and then introduce the reconstruction loss. The pruned model is applicable for different downstream tasks, especially object detection and instance segmentation. 
    
    \item Our result suggests that pixel-level tasks are better than traditional image-level tasks for pruning pre-trained neural networks. Although contrastive learning and classification tasks have been proved to be useful pretraining tasks for large models, the pruning method relying on those tasks may degenerate the transferability. By introducing an appropriate pixel-level task, a pruned model generalizes better on downstream tasks.
    
    \item The proposed method is evaluated on benchmark downstream tasks. It achieves $32.7\%$AP on the COCO dataset when only keeping about $20\%$ parameters of the original model. The superior performance over state-of-the-art result~\cite{girish2021lottery} confirms the effectiveness of our method.
    
\end{itemize}

\section{Related Work}

%Several papers evaluate different pruning criteria, while most pruning methods are based on magnitude pruning, which means removing weight with a small absolute value at each pruning round. In this paper, we use the global magnitude pruning criterion, which means we remove the smallest parameter in the whole network at each pruning round.

\textbf{Pruning and Lottery Tickets Hypothesis}
Pruning aims to remove the unimportant weights of a neural network to reduce computation costs. It was first proposed in \cite{lecun1990optimal} where the authors use the Hessian matrix to estimate the importance of parameters. In \cite{han2015learning}, the authors propose iterative magnitude pruning to achieve a better compression rate. A lot of works follow \cite{han2015learning} setting and achieve promising results. Different to those methods, the lottery tickets hypothesis, proposed in \cite{frankle2018lottery}, suggests that a sparse trainable subnetwork exists in every over-parameterized models. This sparse network can achieve similar performance as the entirty. To verify this assumption, \cite{frankle2018lottery} follow the iterative pruning paradigm but set the model parameter to initial values at each pruning round. Some works \cite{lee2018snip, wang2019picking, tanaka2020pruning} attempt to find the winning tickets in an initialized network. Those methods can find winning tickets in small datasets. However, as stated in \cite{liu2018rethinking}, the original LTH method fails with more complicated datasets and large learning rates. In \cite{renda2019comparing}, the authors find that rewinding the parameter to the early training stage of the neural network, rather than the initial value, can bring profits to winning tickets in complicated datasets. 

%In \cite{morcos2019one}, the authors use the rewind method to confirm that universal tickets exist in classification tasks. \cite{chen2021lottery} use image-level self-supervised learning method to check if they can produce better tickets. In \cite{girish2021lottery} the authors try to apply ImageNet tickets to object recognition tasks. Their result shows that it is hard for ImageNet tickets to transfer on object recognition tasks. 

\textbf{Large Scale Pretrain}
 ImageNet pretraining is widely used in nowadays computer vision training pipeline. It is common sense that using a pretrained network on a large dataset can benefit downstream tasks, both in accuracy and training epochs. 
 %A supervised pretraining ImageNet classifier can generalize to different downstream tasks. 
 Nowadays, self-supervised pretrain methods have become more popular because they can utilize unlabeled data. Image-level self-supervised pretraining has been developed for years \cite{he2020momentum,chen2020simple,grill2020bootstrap,abs-2105-11527}. In those methods, an image is encoded into a single representation vector. The classifier should distinguish a strongly augmented image from irrelevant ones by using their representation vector. Recently, pixel-level or patch-level pretraining has attracted more attention. These methods focus on the recovery of corrupted images \cite{pmlr-v139-ramesh21a, pmlr-v139-touvron21a,he2021masked}. Most of them require an extremely large model such as ViT \cite{dosovitskiy2020image} to achieve better performance. 
 
 \textbf{Autoencoder} is a famous tradional machine learning structure. It is widely used in image denoising \cite{vincent2008extracting} and generative model \cite{kingma2013auto}. Recently, as the image reconstruction task is proposed as a new method in self-supervised pretraining \cite{he2021masked}, autoencoder structure become useful in pretraining task. In order to get abundant semantic information of neural network and avoid cheating model, autoencoder usually use strong regulariziar or data augmentation and should take a lot of time to train. In this paper, we focus on pruning to downstream tasks, rather than get a better autoencoder. By this way, we seperately train the decoder and encoder of our neural network. We also introduce feature map hint method for acceleration. Thus, the decoder only needs to be trained for a much smaller epoch than previous works.
 
 \section{Image-level Tasks are not Sufficient Criterion for Pruning Neural Network}
 \label{sec:Theorem}
In this section, we show the insufficiency of image-level tasks for finding universal tickets. It is common sense that traditional image classification tasks can produce an abundant feature map so that their backbone can transfer to every downstream task. Thus, we use the difference between the original feature map and the pruned version to imply the transferability of a pruned model. We mainly study two simple but important variants of CNN: linear convolutional neural network (LCNN) and one-layer ReLU convolutional neural network (ORCNN). We focus on pruning in LCNN and focus on finetuning in ORCNN. Our results suggest that image-level tasks cannot produce a transferable pruned neural network. We focus on the 1D case in this section, but the 2D case is easy to extend.
\subsection{Preliminary}
{\bf Notations about Tensor} A $k$-th order tensor $\left(a_{i_{1} \ldots i_{k}}\right)$ is a $k$-dimensional array of real numbers $a_{i_{1} \ldots i_{k}}$. We use $||\cdot||$ and $\left< \cdot , \cdot \right> $ to denote the standard $l_2$ norm and inner product of tensors. And we define the normalized $l_2$ distance between tensor $\mathbf{A}$ and $\mathbf{B}$ as 
\begin{equation}
    dist_{l_2}^{N}\left(\mathbf{A},\mathbf{B}\right) = 
    \frac{||\mathbf{A}-\mathbf{B}||}{\sqrt{||\mathbf{A}||||\mathbf{B}||}}
\end{equation}

% {\bf Convolution Tensor} In convolutional neural network(CNN), the convolution tensor is described as follows. Let ${\mathbf{w}}_{ij}=(w_{ij,-s},w_{ij,-s+1},...,w_{ij,s})^{\mathbf{T}} \in \mathbb{R}^{2s+1}(1 \leq i \leq c'$, $1 \leq j \leq c)$ be convolution kernel, where $c'$ is the number of filters and $2s+1$ is the size of convolution kernel. And $\mathbf{W}_{i}=(\mathbf{w}_{i1},\mathbf{w}_{i2},...,\mathbf{w}_{ic})^{\mathbf{T}} \in \mathbb{R}^{c \times (2s+1)}$ denotes $i_{th}$ filter. Then the convolution tensor is defined as a 3-dimensional tensor $\mathbf{W}=(\mathbf{W}_{i})_{1 \leq i \leq c'} \in \mathbb{R}^{c' \times c \times (2s+1)}$.

{\bf Convolution Operator} is a linear operator represented by $*$ . Let $\mathbf{x} = (x_{i,j}) \in \mathbb{R}^{c \times D}$ be the input, where $D$ is the length of input sequence and c is the channel number of input. Let $\mathbf{W} \in \mathbb{R}^{c' \times c \times (2s+1)}$ be the convolution tensor. Then the convolution between $\mathbf{W}$ and $\mathbf{x}$ is defined as:
\begin{equation}
    (\mathbf{W}*\mathbf{x})_{i,j}=\sum_{k=1}^{c}\sum_{l=-s}^{s}w_{ik,l}x_{k,j+l}
\end{equation}
where we use circular padding method, i.e. $x_{i,j+D}:=x_{i,j}$.

{\bf Average Pooling Operator} is also a linear operator $P_{a}$. For any matrix $\mathbf{v}\in \mathbb{R}^{c\times D}$, we have:
\begin{equation}
    P_{a}\mathbf{v}= \frac{1}{D}\sum_{i=1}^{D}\mathbf{v}_{:,i} \in \mathbb{R}^{c}
\end{equation}

% {\bf ReLU Activation Function
% } We use $\sigma(.)$ to denote ReLU activation function, which is defined as $\sigma(x):=\max(0,x)$.

%{\bf $l_2$ Norm of Tensor} Let $\mathbf{A}=(a_{i_1 i_2 ... i_K}) \in \mathbb{R}^{n_1 \times n_2 \times ... \times n_k}$ be a $k$-dimensional tensor, then the $l_2$ norm of $\mathbf{A}$ is defined as:
%\begin{equation}
%    ||\mathbf{A}||_{l_2} = \left(\sum_{i_1 i_2 ... i_k}a_{i_1 i_2 ... i_k}^{2}\right)^{\frac{1}{2}}
%\end{equation}

%{\bf Inner Product of Tensors} Let $\mathbf{A}=(a_{i_1 i_2 ... i_K}),\mathbf{B}=(b_{i_1 i_2 ... i_K}) \in \mathbb{R}^{n_1 \times n_2 \times ... \times n_k}$ be two $k$-dimensional tensors, then the inner product of them is defined as:
%\begin{equation}
%    \left<\mathbf{A},\mathbf{B}\right> = \sum_{i_1 i_2 ... i_k}a_{i_1 i_2 ... i_k}b_{i_1 i_2 ... i_k}
%\end{equation}

%{\bf Normalized $l_2$ Distance of Tensors} Let $\mathbf{A}=(a_{i_1 i_2 ... i_K}),\mathbf{B}=(b_{i_1 i_2 ... i_K}) \in \mathbb{R}^{n_1 \times n_2 \times ... \times n_k}$ be two $k$-dimensional tensors, then the normalized $l_2$ distance of them is defined as:
%\begin{equation}
%    dist_{l_2}^{N}\left(\mathbf{A},\mathbf{B}\right) = 
%    \frac{||\mathbf{A}-\mathbf{B}||_{l_2}}{\sqrt{||\mathbf{A}||_{l_2}||\mathbf{A}||_{l_2}}}
%\end{equation}

%\subsection{Two Simplified Variants of CNN}

%Now, we introduce two simple but important variants of CNN:\\ %linear convolutional neural network (LCNN) and one-layer ReLU convolutional neural network (ORCNN).
{\bf Linear Convolutional Neural Networks (LCNN):} LCNN is a linear mapping $\mathcal{C}:\mathbb{R}^{c\times D}\rightarrow \mathbb{R}^{m_{L+1}\times D}$, which can be defined as 
    \begin{equation}
    \label{eq:LCNN}
        \mathcal{C}(x) \coloneqq W^{L}*W^{L-1}*...W^0*x
    \end{equation}
    where $x\in \mathbb{R}^{c\times D}$,$W^l\in\mathbb{R}^{m_{l+1}\times m_{l}\times (2s+1)}$. %$D$ is the number of input sequence length,$c$ is the channel number of input. $m_l$ is the channel number of $l$-th layer, $m_0=c$. $2s+1$ is the kernel size. $*$ is the convolution operator. 

{\bf One-hidden-layer ReLU Convolutional Neural Networks (ORCNN):} Let $\mathbf{x} = (x_{i,j}) \in \mathbb{R}^{c \times D}$ be the input. %where $c$ is the number of input channels and $D$ is the length of input sequence. 
$\mathbf{W} \in \mathbb{R}^{m \times c \times (2s+1) 
    }$ is the convolution tensor, where $m$ is the channel number of feature maps. ORCNN is defined as the following:
    \begin{equation}
    \begin{aligned}
        \mathbf{x}^{conv} &= \frac{1}{\sqrt{m}}\sigma(\mathbf{W}*\mathbf{x})
        \\
        \mathbf{x}^{pool} &= P_a\mathbf{x}^{conv}
        \\
        f(\mathbf{x}) &= \left<\mathbf{a},\mathbf{x}^{pool}\right>
    \end{aligned}
    \end{equation}
    where $\mathbf{x}^{conv}$ , $\mathbf{x}^{pool}$ are hidden-layer outputs and $f(\mathbf{x})$ is the predicted label. We use $f_{\mathbf{a},\mathbf{W}}(\mathbf{x})$ to denote the whole newtork. $\sigma(\cdot)$ denotes ReLU activation function which is defined as $\sigma(\cdot):=\max(\cdot,0)$. 
    %$P_a$ denotes average pooling operator which is also defined in preliminary section. 
    $\mathbf{a}=(a_1,a_2,...,a_m)^{\mathbf{T}} \in \mathbb{R}^{m}$ are the fully connected weights.
% \begin{itemize}
%     \item 
%     \item 
% \end{itemize}

\subsection{Pruning in LCNN}

 \label{sc:Pruning in LCNN}
 
In this section, we investigate the pruning step in LCNN. We first claim a simple proposition of LCNN:

{\bf Claim} Let $\mathcal{C}$ be an LCNN, there must exists another LCNN $C'$, such that
\begin{equation}
    P_{a}\mathcal{C}(x)= P_{a}\mathcal{C'}(x), \mathcal{C}(x)\neq\mathcal{C'}(x)
\end{equation}

This is a direct result by considering the translation symmetry in LCNN. Now, the question becomes ``Can this LCNN $\mathcal{C'}$ be found by pruning algorithm?". At least, can we find an LCNN $\mathcal {C}'$ by pruning, such that the change of $||P_{a}\mathcal{C}(x)- P_{a}\mathcal{C'}(x)||$ is small while $ ||\mathcal{C}(x)-\mathcal{C'}(x)||$  is large? To answer this question, we come out the following theorem: 
\begin{theorem}
\label{th:prune LCNN}
For any random initialized LCNN, where parameter is initialized as i.i.d $\mathcal{N}(0,\Delta)$. Then, for any $p < 0.11$, we can prune $ p $ proportion of weights and get a new LCNN $\mathcal{C'}$ with high probability, such that:
\[\frac{||P_a\mathcal{C}(x)-P_a\mathcal{C'}(x)||}{||P_a\mathcal{C}(x)||} < C_1 p^{3/2}\]
\[\frac{||\mathcal{C}(x)-\mathcal{C'}(x)||}{||\mathcal{C}(x)||} > C_2 p^{1/2}\]
Here $C_1$ and $C_2$ are constants related to the kernel size $s$ and the depth $L$
\end{theorem}
This theorem means that if we initialize the LCNN properly, we can find some neurons such that removing those neurons does not change the image-level feature vector a lot but destroys the feature map structure. Thus, using a pruning criterion based on image-level loss can not preserve the feature map of LCNN. The detailed proof is in Appendix \ref{appendix:LCNN}.

\subsection{Finetuning in ORCNN}

In this section, we focus on finetuning in ORCNN. Let $f_{\mathbf{a},\mathbf{W}}(\mathbf{x})$ denote the ORCNN parameterized by fully connected weight $\mathbf{a}$ and convolution tensor $\mathbf{W}$.
%First, we describe three phases of pruning process (pre-trained phase, pruning phase and finetuning phase). Furthermore we demonstrate the insufficiency of image-level tasks for finding universal tickets by analyzing the optimization process in finetuning phase.
% We use $f_{\mathbf{a},\mathbf{W}}(\mathbf{x})$ to denote the output in input $\mathbf{x}$ of the ORCNN parameterized by fully connected weight $\mathbf{a}$ and convolution tensor $\mathbf{W}$.
The pruning pipeline can be formalized as the following three phases:

\begin{itemize}
    \item {\bf Pre-trained Phase:} We randomly initialize the parameters $\mathbf{a}$ and $ \mathbf{W}$ to $\mathbf{a}_0$ and $ \mathbf{W}_0$. Then, we train the model via image-level tasks on the given labeled dataset $S$ and derive a pre-trained model $f_{\mathbf{a}_{pre},\mathbf{W}_{pre}}(\mathbf{x})$. 
    \item {\bf Pruning Phase:} We apply the structured pruning method to ORCNN with the pruning rate $p$.
    \item {\bf Finetuning Phase:} We first reset unpruned parameters to initial values $\mathbf{a}_0$ and $ \mathbf{W}_0$. Next, we finetune the network parameters on the same labeled dataset $S$ via the gradient descent algorithm. Finally, we derive the finetuned model $f_{\mathbf{a}_{fin},\mathbf{W}_{fin}}(\mathbf{x})$.
\end{itemize}

We mainly consider the training process in the finetuning phase. In the finetuning phase, we use the same dataset $S=\{ (\mathbf{x_1},y_1),(\mathbf{x_2},y_2),...,(\mathbf{x_n},y_n)\}$ as in the pretraining phase and use loss function $ L(f):=\frac{1}{2}\sum_{i=1}^{n}(f(\mathbf{x_i})-y_i)^{2}$. We use the gradient descent method to update the convolutional tensor $\mathbf{W}$ and freeze the fully connected weights $\mathbf{a}$:

\begin{equation}
    \mathbf{W}(t+1) = \mathbf{W}(t) - \eta \frac{\partial L}{\partial \mathbf{W}(t)}
\end{equation}

% More details of finetuning phase are as follows.

% {\bf DataSet} In finetuning process, we use a given labeled dataset $S=\{ (\mathbf{x_1},y_1),(\mathbf{x_2},y_2),...,(\mathbf{x_n},y_n)\}$, where $\mathbf{x_i} \in \mathbb{R}^{c \times d}$ is input and $y_i \in \mathbb{R}$ is real-value label.

% {\bf Quadratic Loss Function} We focus on the empirical risk minimization problem with a quadratic loss, we want to minimize
% \begin{equation}
%     L(f):=\frac{1}{2}\sum_{i=1}^{n}(f(\mathbf{x_i})-y_i)^{2}
% \end{equation}
% where $f$ is the model prediction function.

% {\bf Gradient Descent Algorithm} In the optimization process, we fix the fully connected weight $\mathbf{a}$ and apply gradient descent to optimize the convolutional tensor $\mathbf{W}$.
% \begin{equation}
%     \mathbf{W}(t+1) = \mathbf{W}(t) - \eta \frac{\partial L}{\partial \mathbf{W}(t)}
% \end{equation}
where $\eta$ is the learning rate, and $t$ denotes $t^{th}$-iter.

As the original pre-trained model has a strong transferability to diverse downstream tasks, we believe the pre-trained model can learn a good representation of the data. Specifically, the pre-trained ORCNN can derive the feature maps from the image data by the pre-trained convolution tensor $\mathbf{W}_{pre}$.
Thus, the `difference' between convolution tensors $\mathbf{W}_{fin}$ and $\mathbf{W}_{pre}$ implies the transferability of pruned model.

%Thus we focus on the finetuned convolution tensor $\mathbf{W}_{fin}$, which represents the finetuned model's ability to extract important features and is expected to close to that in the original model such that the pruned model can also extract information well. 

In order to measure the difference between convolution tensors, we multiply an arbitrary rotation operator $\mathbf{Q}$ on the $\mathbf{W}_{fin}$ to recover its density. Then, we calculate the minimal normalized $l_2$ distance between $\mathbf{Q}\mathbf{W}_{fin}$ and $\mathbf{W}_{pre}$. The following Theorem \ref{th:finetune OR} characterizes the lower bound of the distance under the over-parameterized setting.

\begin{theorem}
\label{th:finetune OR}
Assume that we set the channel number of feature maps $m=\Omega(\frac{1}{{\delta}^{2}}poly(n))$, and the finetuning learning rate $\eta$ is sufficiently small. After the finetuning phase, the finetuned convolution tensor is $\mathbf{W}_{fin}$. Then with probability at least
$1-\delta$ over the random initialization in the pre-trained phase, we have
\begin{equation}
    \min_{\mathbf{Q} \in \mathbb{O}} \{ dist_{l_2}^{N}\left(\mathbf{Q}\mathbf{W}_{fin}, \mathbf{W}_{pre}\right) \} \geq \frac{p}{2}
\end{equation}
where $\mathbb{O}$ is rotation operator space and $p$ is the pruning rate.
\end{theorem}

The main idea of the proof is to analyze the dynamics of the model Gram matrix in the gradient descent process. The detailed proof can be found in Appendix \ref{appendix:ORCNN}. 

Theorem \ref{th:finetune OR} suggests the lower bound of normalized $l_2$ distance between them is growing linearly with respect to the pruning rate $p$. It reveals the pruned model's ability to extract features is less than the original although it may have the same good performance as the original model in image-level tasks. Therefore, we demonstrate the insufficiency of image-level tasks for finding universal tickets. 

  \begin{figure}[ht]
\vskip 0.05in
\begin{center}
\centerline{\includegraphics[width=\textwidth]{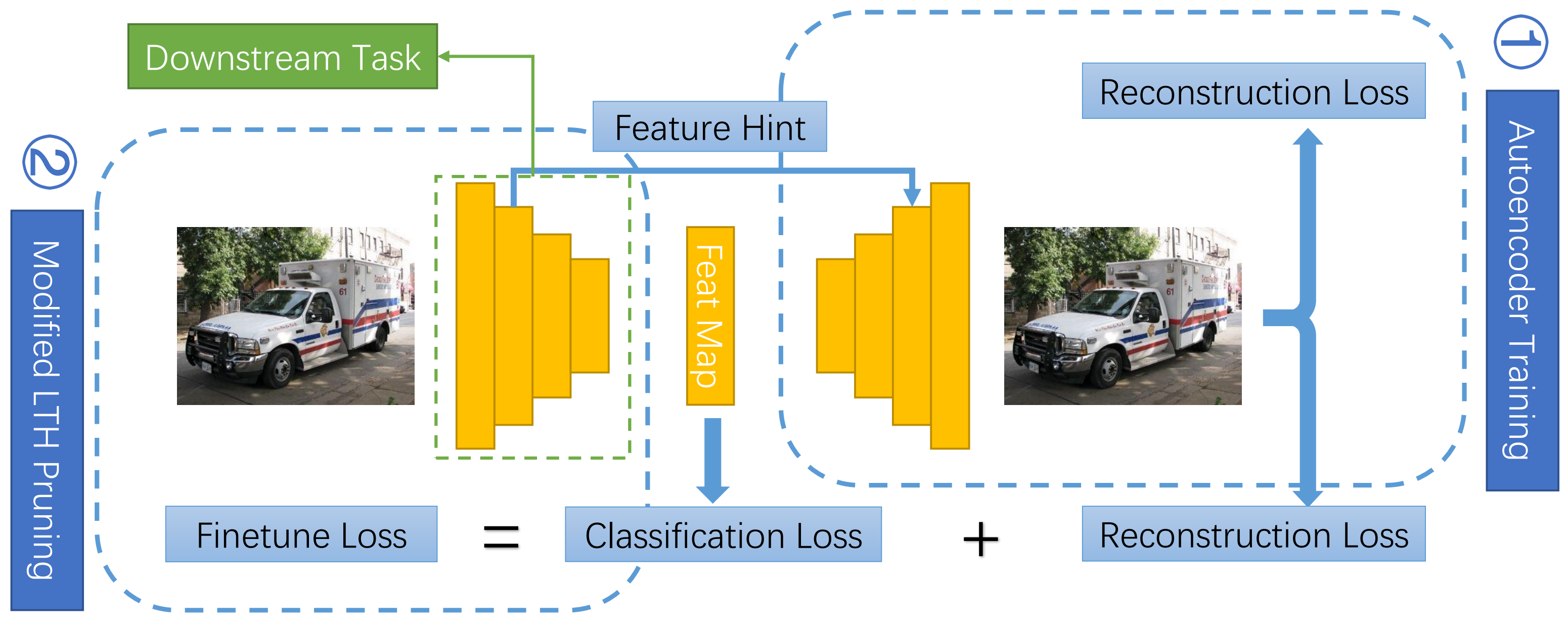}}
\caption{{\bf Overview of our framework.} First, we use the reconstruction loss and the feature map hint method to train an autoencoder structure. Next, we use the modified LTH algorithm for pruning the neural network. We combine classification loss and image reconstruction loss for finetuning procedure. Then, we transfer the encoder part to the downstream tasks. The encoder part is frozen during the autoencoder training process, while the decoder is frozen in the Modified LTH pruning process.}
\label{fig:structure}
\end{center}
\vskip -0.05in
\end{figure}
\section{Method}
As we discussed in Section 3, only focusing on the image-level task during the pruning procedure will lead to a degenerated feature map. In general, using image-level loss as a pruning criterion tends to remove image details and destroy the structure of the feature map. That untraceable information will cause a significant accuracy drop on detection or segmentation tasks. To mitigate this problem, we attempt to introduce the pixel-level task to the traditional pruning framework. Our framework can be formalized into three stages: 

$i$) Train an autoencoder. We modify the pretrained model to the encoder of an autoencoder structure. The last feature map of the original model becomes the compressed code of autoencoder; then, we freeze the encoder and start training. We use the feature map hint method (illustrated in Section \ref{sec:autoencoder training}) to accelerate the training process and improve performance. 

$ii$) Prune the encoder. After decoder training, we prune the encoder part with reconstruction loss and classification loss simultaneously. During this pruning step, the decoder is frozen to keep the information gathered from the encoder. We follow a modified LTH pruning pipeline to get better performance. 

$iii$) Adapt the encoder to the downstream tasks according to the standard transfer learning setting. 

 The overall structure design is described in Figure \ref{fig:structure}. We will describe the first two steps in our framework in the following sections.
%The first convolution layer and the max-pooling layer of resnet50 adjust to the proper shape of downstream datasets. \warn{(this part will not influence the final performance but make the transferring step faster. The compression rate change is negligible because the first kernel only contains 0.03\% parameters of ResNet50)}.

\subsection{Autoencoder Training}
\label{sec:autoencoder training}

Image reconstruction is a conventional computer vision task but was introduced as a pretraining method recently\cite{he2021masked}. Autoencoder is the basic architecture of image reconstruction. As the first step of our framework, the original model will be embedded in the autoencoder structure, which will be trained until the decoder captures the pixel-level information.

In this paper, we focus on ResNet structure, but our method can easily generalize to other kinds of structures. We remove the last pooling layer and fully connection layer of the original model as the encoder part. In this way, the final feature map is regarded as the compressed code of the autoencoder. Different from unsupervised pretraining, the decoder part in our method is an inversed ResNet. The training purpose of an autoencoder is to minimize 
\begin{equation}
    \label{eq:recloss}
    \mathcal{L}_{rec} = \frac{1}{N}\sum_{i=1}^{N} ||\mathcal{D}(\mathcal{F}(x_i))-x_i||^2
\end{equation}
where $N$ is the number of training samples, $\mathcal{F}$ represents the encoder part, $\mathcal{D}$ represents the decoder part, $x_i$ is the input image.  Obviously, without any constraint on $\mathcal{F}$ or $\mathcal{D}$, the loss function will lead to a trivial solution. Therefore, we freeze the parameters in $F$ during the autoencoder training process.

\subsection{Feature Map Hint}

In previous works, autoencoder is widely used in generative \cite{pmlr-v139-ramesh21a} and reconstruction \cite{gondara2016medical} tasks. Such tasks require a lot of training time because they need to finetune both the encoder and decoder. Unlike those training strategies, we treat the feature map as an effective representation and should not change during the autoencoder training step. Although the feature map has abundant information, the recovery of the whole image is difficult due to the resolution restriction.

We use the previous stage's feature map as a hint to the inversed ResNet decoder for better image reconstruction results. Directly transporting the feature map to the decoder part must lead to a fault model. The decoder tends to rely on low-level features while ignoring the high stage's information. Therefore, we mix the original feature map and the following decoder's feature map with a specific proportion $t$. Formally, let $f_i$ be the feature map in the encoder stage-$i$, $g_i$ is the output feature in the decoder stage-$i$, the input feature map of the decoder $i+1$ stage is:
\begin{equation}
\label{eq:hint}
    g'_i=(1-t)g_i + tf_i
\end{equation}
This mixing method can stabilize the finetuning process and avoid fault models. In practice, the proportion $t$ will be a relatively smaller number. In Section \ref{sec:ablation feat map}, we conduct an ablation study on the choice of feature map. The final result shows that using $f_3$ and $f_4$ as feature map hints for the decoder can achieve the best performance. 

\subsection{Reconstruction Loss in Pruning Step}

With the autoencoder structure, we can add the reconstruction loss to the pruning process. The loss function during the pruning process is composed of two parts: $L_{class}$ refers to the traditional classification loss, and $L_{rec}$ refers to the reconstruction loss. We use a hyperparameter $\lambda$ to balance between $\mathcal{L}_{class}$ and $\mathcal{L}_{rec}$ 
\begin{equation}
\label{eq:finetuneloss}
    \mathcal{L} = \mathcal{L}_{class}(\mathcal{F}) + \lambda \mathcal{L}_{rec}(\mathcal{F},\mathcal{D})
\end{equation}
Notice that the decoder $\mathcal{D}$ is related to $\mathcal{L}_{rec}$ in the above function, which means the decoder will be trained during the finetuning process. However, we hope the decoder part guides the finetuning process and transfers pixel-level information to the encoder. The decoder change will perturb the information that remains in the decoder and may lead to an unexpected result. It also slows down the training. Therefore, we freeze the parameter in the decoder part during the pruning process.

\subsection{Modified LTH Pruning}
\label{sec:modified LTH}

There are two widely used pruning pipelines, IMP and LTH, where LTH reset the parameter to the initial value, but IMP does not. In this section, We argue that neither setting is the most capable method to find universal tickets. We introduce a modified LTH pipeline, which is showed more powerful to find universal tickets.

In previous works, \cite{chen2021lottery} used an IMP method to produce universal tickets. Although the IMP method usually provides better accuracy on the upstream tasks, it may cause the neural network to fall in the minima of the upstream task while it is hard to finetune on the downstream task. In \cite{girish2021lottery}, the authors examine whether the ImageNet tickets produced by LTH can work for detection tasks. We should point out that their method does not utilize the pretrained network power and limits the performance of their method. We try to combine the strengths of 
those two methods.

 %Compared to directly applying LTH, a slightly different but useful algorithm was introduced in \cite{renda2019comparing}. The authors reset parameters to some early values rather than initial values. They show that this method can stabilize the relation between tickets and their final performance on a large dataset, such as ImageNet. Therefore, their method can find a better ticket than the original LTH algorithm. In \cite{chen2021lottery}, the authors argue that a pretrained method is a special initialization method. Following those inspirations, we use the pretrained value of the whole network as the early trained parameters. 

We describe our modified LTH algorithm in detail. Inspired by \cite{chen2021lottery}, we replace the initial values in the original LTH method with the pretrained values. Let $f(x;\theta)$ represent the neural network, $\theta \in \mathbb{R}^n$. Pruning some parameters means we permanently set some dimensions of $\theta$ to zero. In this way, we can use mask $m\in \{0,1\}^{n}$ to describe the pruned parameter. Thus, a pruned network can be described by $f(x;m\odot\theta)$, where $m_i=0$ means we prune this parameter from the whole neural network. Let $\theta_{pre}$ represent the pretrained value of the original neural network. We train the network for $T$ epochs and prune the smallest $p$ proportion of unmasked parameters for every training round. After the prune step, the parameters $\theta$ reset to the pretrained values $\theta_{pre}$. Then another training round begins. In Algorithm \ref{alg:Modified LTH}, we state the pseudo-code of our method.

 \begin{algorithm}[tb]
  \caption{Modified LTH}
  \label{alg:Modified LTH}
\begin{algorithmic}
  \STATE {\bfseries Input} A neural network $f(x;\theta)$, pretrained value $\theta_{pre}$, parameter remain percentage at each round $r=1-p$, total pruning round $R$, finetuning strategy $s$ in each round 
  \STATE {\bfseries Output} A set $S$ of pruned tickets at different pruning levels, $S = \{(1-r^i,f_i)|1 \leq i \leq R\}$
  \STATE $m\leftarrow \mathbbm{1}$, $\theta \leftarrow \theta_{pre}, S\leftarrow \emptyset$
  \FOR{$i=1$ {\bfseries to} $R$}
  \STATE Finetune $f(x;\theta\odot m)$ following the strategy $s$
  \STATE Prune smallest non-masked $p$ values, update $m$
  \STATE $\theta \leftarrow \theta_{pre}$
  \STATE $S\leftarrow S\cup \{(1-r^{i},f(x;\theta\odot m)\}$
  \ENDFOR
\end{algorithmic}

\end{algorithm}
\vspace{-10pt}
% \begin{algorithm}[tb]
%   \caption{Bubble Sort}
%   \label{alg:example}
% \begin{algorithmic}
%   \STATE {\bfseries Input:} data $x_i$, size $m$
%   \REPEAT
%   \STATE Initialize $noChange = true$.
%   \FOR{$i=1$ {\bfseries to} $m-1$}
%   \IF{$x_i > x_{i+1}$}
%   \STATE Swap $x_i$ and $x_{i+1}$
%   \STATE $noChange = false$
%   \ENDIF
%   \ENDFOR
%   \UNTIL{$noChange$ is $true$}
% \end{algorithmic}
% \end{algorithm}
  \begin{figure*}[ht]
\vskip 0.05in
\begin{center}
\begin{minipage}{0.5\textwidth}
\includegraphics[width=\linewidth,trim={0 0 0 0.7cm},clip]{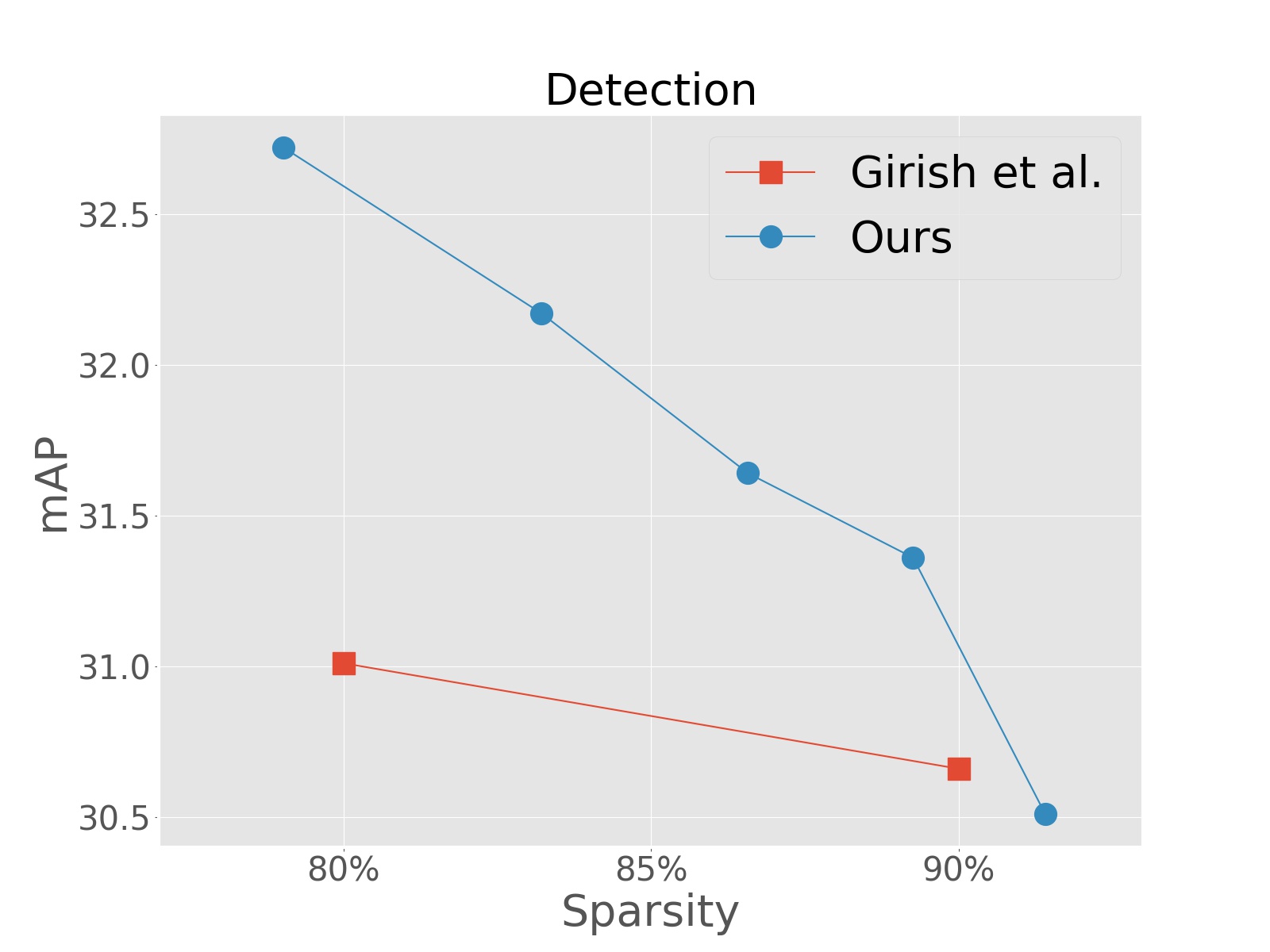}
\end{minipage}%
\begin{minipage}{0.5\textwidth}
\includegraphics[width=\linewidth,trim={0 0 0 0.7cm},clip]{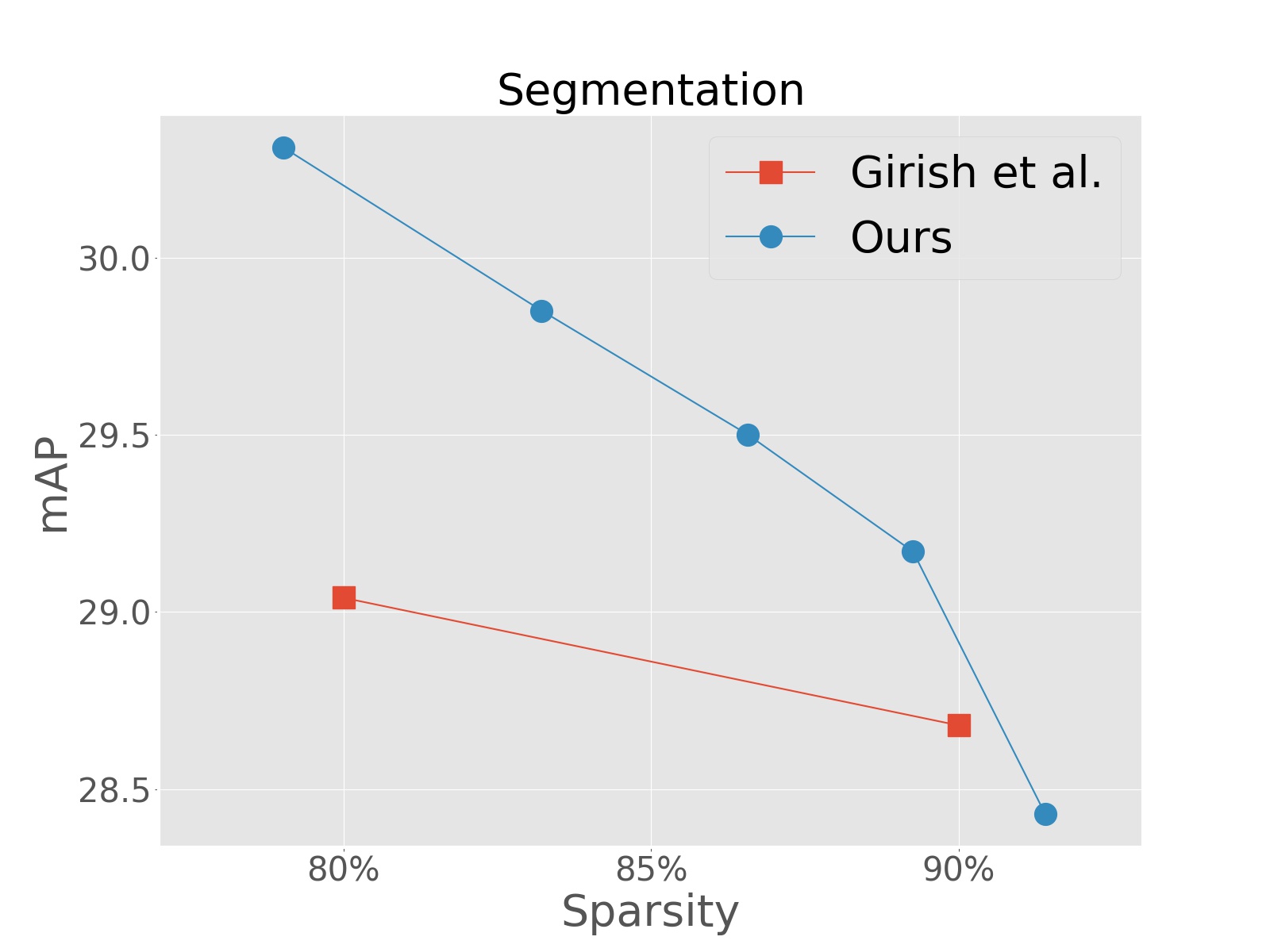}
\end{minipage}
% \centerline{\includegraphics[width=0.8\columnwidth]{detection.jpg}}
\caption{{\bf Performance on the COCO detection and segmentation dataset}. It is noticeable that our method outperforms the reported result in  \cite{girish2021lottery}. We can produce transferable tickets at sparsity around $80\%$. }
\label{fig:detection and segmentation result}
\end{center}
\vskip -0.05in
\end{figure*}

\section{Experiment}
\subsection{Setting}

\textbf{Dataset}
For the autoencoder training step and modified LTH pruning step, we conduct all experiments on ImageNet \cite{deng2009imagenet}. For the image-level transfer task, we still focus on image classification. We evaluate the tickets gathered from the ImageNet dataset on Cifar10 \cite{krizhevsky2009learning}, Cifar100\cite{krizhevsky2009learning}, and SVHN \cite{netzer2011reading}. We also show the accuracy of ImageNet. For pixel/patch-level task transfer, we investigate object detection and instance segmentation. As stated in \cite{girish2021lottery}, ImageNet tickets transfer to the COCO \cite{lin2014microsoft} dataset is harder than transfer to small detection dataset such as VOC datasets. Therefore, We use the COCO dataset as a standard benchmark.

\textbf{Model}
We evaluate our method with ResNet50, a standard CNN model on the ImageNet and COCO object detection/instance segmentation task backbone. We use official PyTorch pretrained weights as our pretrained values. The decoder is an inverse ResNet architecture. This decoder part is removed in the downstream tasks, and only the encoder part transfer. For classification transfer tasks, we adjust the first kernel size of ResNet50 to $3\times3$ and remove the first max-pooling layer. We use a famous structure mask RCNN\cite{he2017mask} as the segmentation and detection head for object detection and instance segmentation tasks. 

% We follow the standard $1\times$ FPN training configs defined in detectron2.

\textbf{Training and Pruning Setting} 
In the autoencoder training step, we train the decoder with the AdamW optimizer. We use a multi-step learning rate schedule with an initial learning rate 1e-4 and $\times0.1$ at the 10, 30 epoch. The total training epoch is 50; the batch size is 512; weight decay is 2e-4.
In the modified LTH pruning step, We follow the pruning setting in \cite{chen2021lottery}, where for each pruning round, we prune 20\% parameters. Each pruning round has 10 epochs. We use the SGD optimizer, and the learning rate is kept 3e-4, the batch size is 512, momentum is 0.9. The reconstruction penalty $\lambda$ defined in (\ref{eq:recloss}) is 10, the feature map hint proportion is 0.1.
For classification task transfer setting, we follow the setting in \cite{chen2021lottery}. We draw the accuracy-compression rate curve for different classification tasks. For segmentation and detection tasks, we use a standard Detectron2\cite{wu2019detectron2} FPN $1\times$ training config. We mainly evaluate the performance at 7,8,9,10,11 pruning round (79.03\%,83.22\%,86.58\%,89.26\%,91.41\%) to make a comparison with result in \cite{girish2021lottery}. The numbers in the brackets represent the sparsity at each pruning round.

\subsection{Detection and Segmentation Results}

%   \begin{figure}[ht]
% \vskip 0.1in
% \begin{center}
% \centerline{\includegraphics[width=0.8\columnwidth]{segmentation.jpg}}
% \caption{{\bf The result of coco segmentation}.}
% \label{fig:detection result}
% \end{center}
% \vskip -0.1in
% \end{figure}
We evaluate our method on the COCO dataset. In Figure \ref{fig:detection and segmentation result}, we compare our method with the reported ImageNet tickets results in \cite{girish2021lottery}. Notice that our results surpass the baseline results at every point of the accuracy-sparsity curve. Our method can achieve 32.7 mAP on the detection task and 30.3 mAP on the segmentation task at sparsity 79.03\%. To make a fair comparison with the reported result in \cite{girish2021lottery}, we interpolate our accuracy-sparsity curve at 80\% and 90\% sparsity. The result is listed in Table \ref{tab:interpolate result}. At high-level sparsity, e.g., 92\% sparsity, our method's performance goes down due to the limitation of ResNet50 itself. As stated in \cite{girish2021lottery}, it is hard for ResNet50 to find winning tickets at 90\% sparsity, even in the setting of directly applying LTH on the downstream task.

\subsection{Classification Results}
  \begin{figure*}[ht]
\vskip 0.05in
\begin{center}
\begin{minipage}{.5\textwidth}
\includegraphics[width=\linewidth,trim={0 0 0 0.7cm},clip]{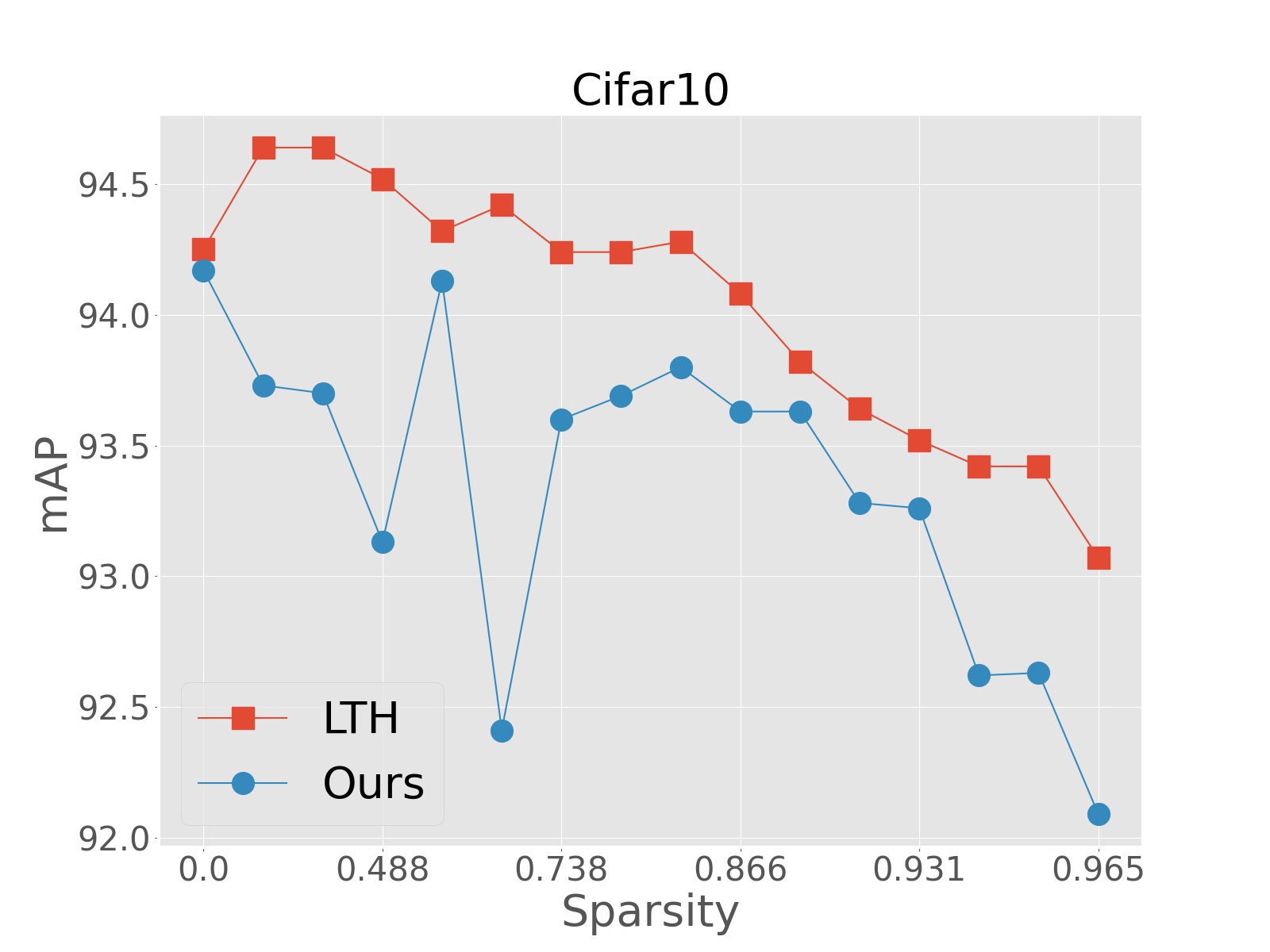}
\end{minipage}%
\begin{minipage}{0.5\textwidth}
\includegraphics[width=\linewidth,trim={0 0 0 0.7cm},clip]{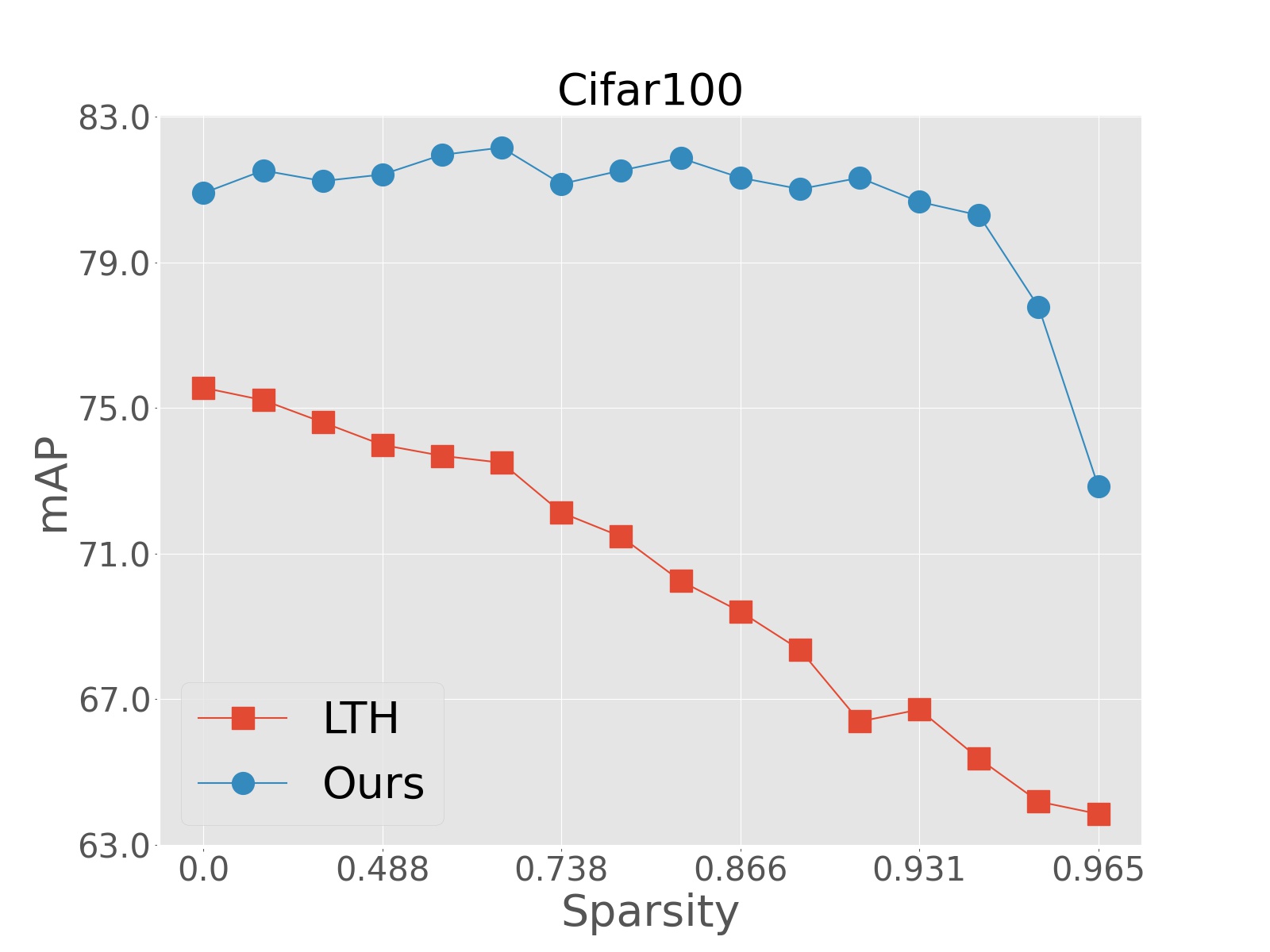}
\end{minipage}\\
\begin{minipage}{0.5\textwidth}
\includegraphics[width=\linewidth,trim={0 0 0 0.7cm},clip]{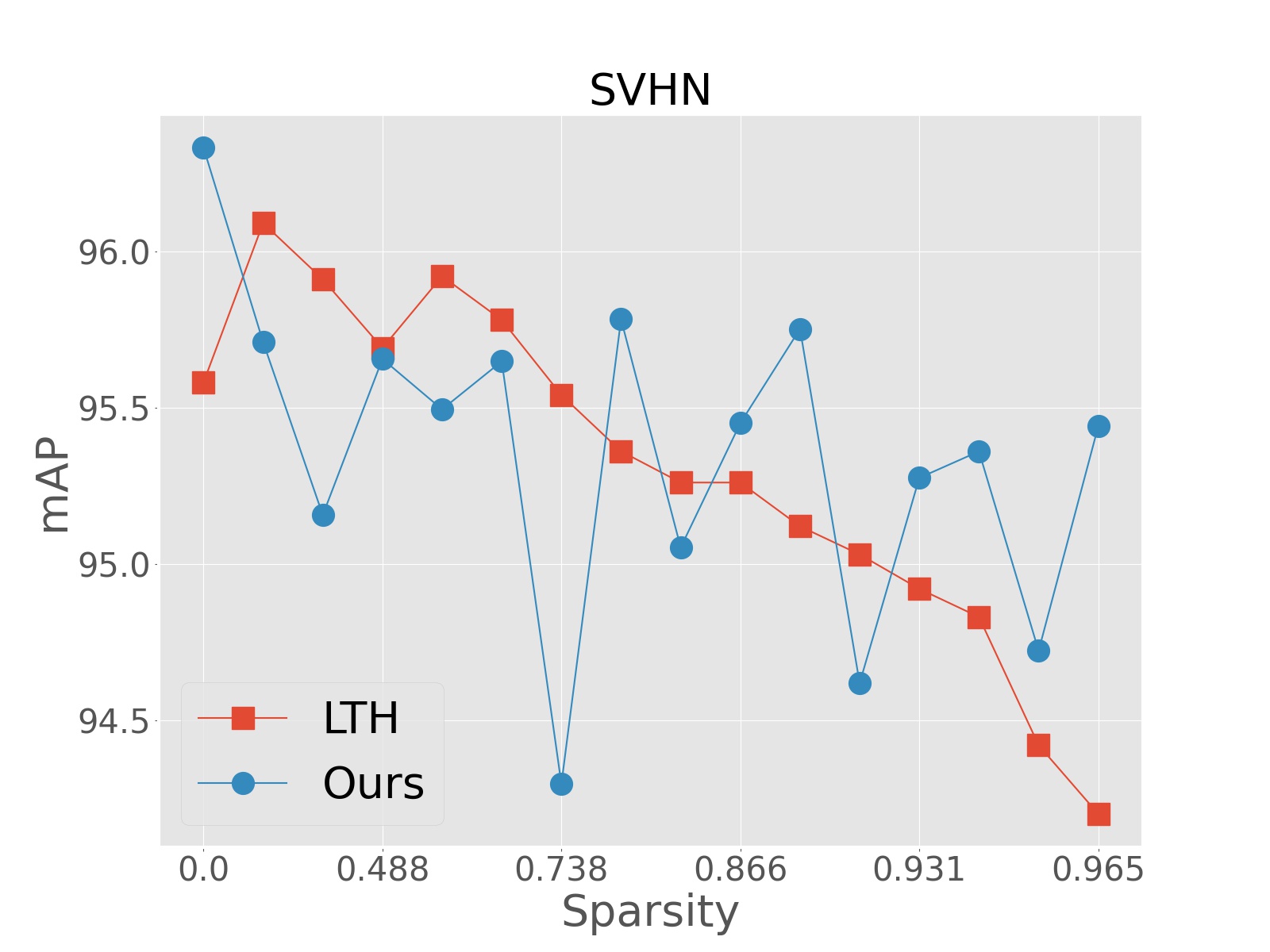}
\end{minipage}
% \centerline{\includegraphics[width=0.8\columnwidth]{detection.jpg}}
\caption{{\bf Classification transfer results.} From left to right are the results of Cifar10, Cifar100, SVHN. In Cifar10 and SVHN datasets, our method is comparable to the directly applying LTH method. At Cifar100, our method outperforms with LTH method, showing the transferability of our selected tickets.}
\label{fig:classification transfer result}
\end{center}
\vskip -0.05in
\end{figure*}
{\bf Baseline setting} We compare our results with the baseline that directly performs LTH on the target dataset in classification task transfer. We perform 15 pruning rounds. Each pruning round has 182 epochs and will cut 20\% parameters. The learning rate starts from 0.1 in each round, $\times 0.1$ at 91, 136. The weight decay is 2e-4. This setting is the same as the task setting in \cite{chen2021lottery}. 

We evaluate our tickets on Cifar10, Cifar100, SVHN. The results are presented in Figure \ref{fig:classification transfer result}. Our method is comparable or even better than directly applying the LTH method on the target dataset, especially in Cifar100 datasets. In Cifar100, our method achieves over $80\%$ accuracy, while the LTH method only achieves 76\% at the beginning of training.

\begin{figure}[htb]
% \vskip 0.1in
\begin{center}
\centerline{\includegraphics[width=0.5\textwidth]{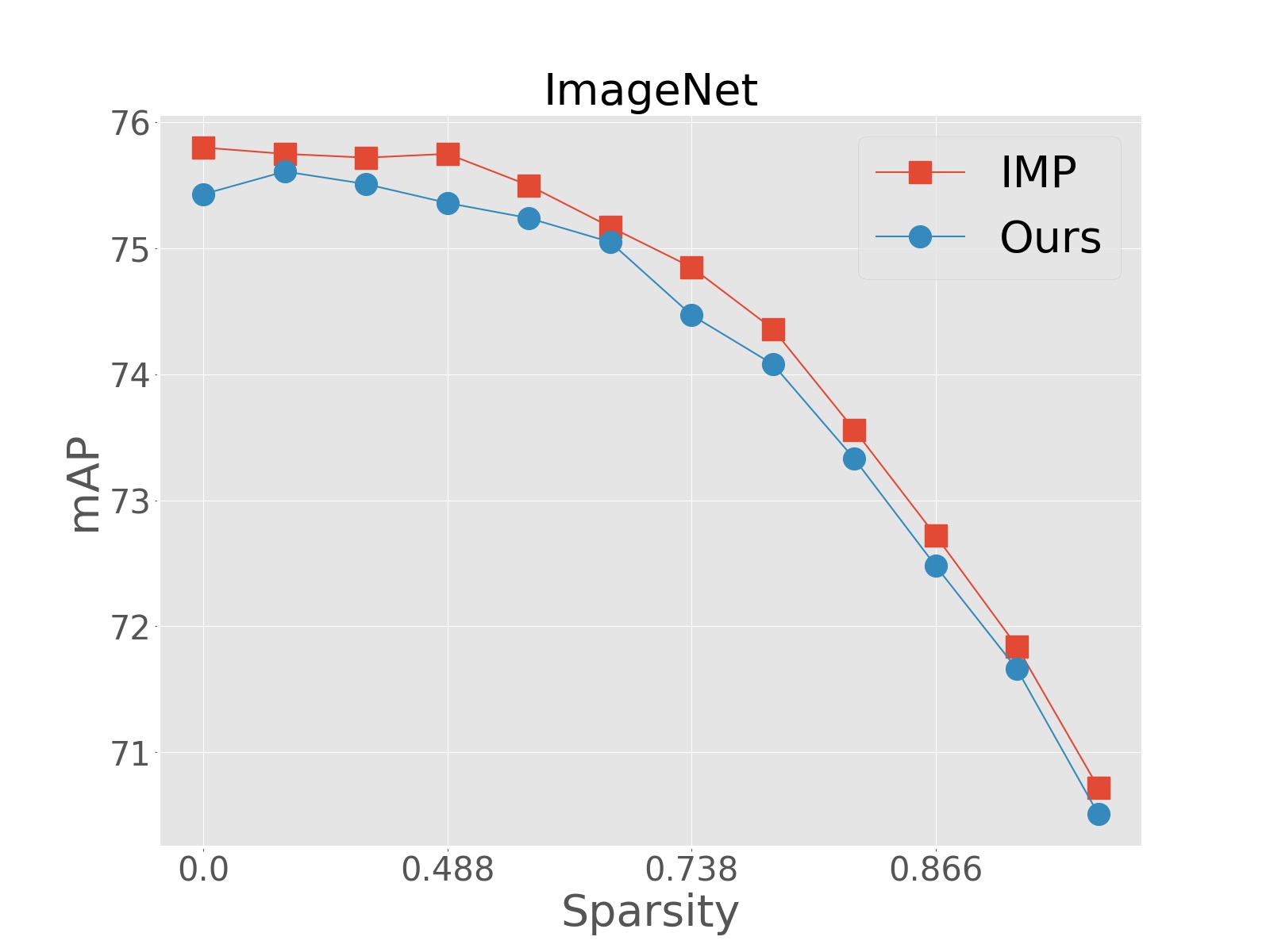}}
\caption{{\bf The comparison between our method and IMP on ImageNet}. We find that our method has a similar performance as the IMP result. Considering that our method only finetunes 10 epochs for each ticket, we believe that our ticket is also winning tickets on ImageNet.}

\label{fig:imagenet}
\end{center}
\vskip -0.1in
\end{figure}

Although we are mainly concerned about the pruned network's transferability, our method still achieves comparable ImageNet unstructured pruning results. As shown in Figure \ref{fig:imagenet}, our method has a similar performance with the iterative magnitude unstructured pruning result, which implies that our tickets are also winning tickets on the ImageNet.

\section{Ablation Study}
\begin{table}[t]
\caption{Object Detection and Segmentation Results. (mAP)}
\label{tab:interpolate result}
\vskip 0.05in
\begin{center}
\begin{small}
%\begin{sc}

\begin{tabular}{|c|cc|cc|}
\hline
Task                           & \multicolumn{2}{c|}{Detection}                         & \multicolumn{2}{c|}{Segmentation}                      \\ \hline
\multicolumn{1}{|c|}{Sparsity} & \multicolumn{1}{c|}{80\%}  & \multicolumn{1}{c|}{90\%} & \multicolumn{1}{c|}{80\%}  & \multicolumn{1}{c|}{90\%} \\ \hline
\cite{girish2021lottery}                           & \multicolumn{1}{c|}{31.0} & 30.7                     & \multicolumn{1}{c|}{29.0} & 28.6                     \\ \hline
ours (interpolated)                         & \multicolumn{1}{c|}{32.6} & 31.1                     & \multicolumn{1}{c|}{30.2} & 28.9                     \\ \hline

\end{tabular}

%\end{sc}
\end{small}
\end{center}
\vskip -0.2in
\end{table}
\begin{table}[t]

\caption{Comparison with IMP Method. Object Detection and Segmentation Results. (mAP)}
\label{tab:IMPvsLTH}
\vskip 0.05in
\begin{center}
\begin{small}
%\begin{sc}

\begin{tabular}{|c|cc|cc|}
\hline
\multirow{2}{*}{\begin{tabular}[c]{@{}c@{}}Sparsity\\ (backbone)\end{tabular}} & \multicolumn{2}{c|}{Detection} & \multicolumn{2}{c|}{Segmentation} \\ \cline{2-5} 
                            & \multicolumn{1}{c|}{IMP} & Ours & \multicolumn{1}{c|}{IMP}   & Ours  \\ \hline
                         79.02\%   & \multicolumn{1}{c|}{32.1}    &  32.7   & \multicolumn{1}{c|}{29.9}      &   30.3   \\
                         83.22\%  & \multicolumn{1}{c|}{31.8}    &  32.3   & \multicolumn{1}{c|}{29.6}      &    29.8  \\ 
                        86.57\%  & \multicolumn{1}{c|}{31.3}    &    31.6 & \multicolumn{1}{c|}{29.1}      &   29.5   \\
                        89.26\%  & \multicolumn{1}{c|}{30.8}    &   31.4  & \multicolumn{1}{c|}{28.7}      &   29.2   \\
                        91.41\%  & \multicolumn{1}{c|}{30.1}    &    30.5 & \multicolumn{1}{c|}{28.0}      &    28.4  \\\hline
\end{tabular}

%\end{sc}
\end{small}
\end{center}
\vskip -0.1in
\end{table}

\subsection{Comparison vs IMP Method}

As we argued in Section \ref{sec:modified LTH}, the IMP method is not the best way to create universal tickets. In this section, we compare the transferability of tickets produced by IMP and our method. We evaluate those tickets on detection downstream tasks at different sparsity. It is easy for the IMP method to get a more sparse network in complicated datasets. However, we find out that in the transferability tasks, our method brings significant AP improvement in the downstream detection and segmentation datasets. The results are shown in Table \ref{tab:IMPvsLTH}

\subsection{Feature Map Hint Selection}
\label{sec:ablation feat map}
  \begin{figure}[htb]
\vskip 0.1in
\begin{center}
\centerline{\includegraphics[width=0.5\textwidth]{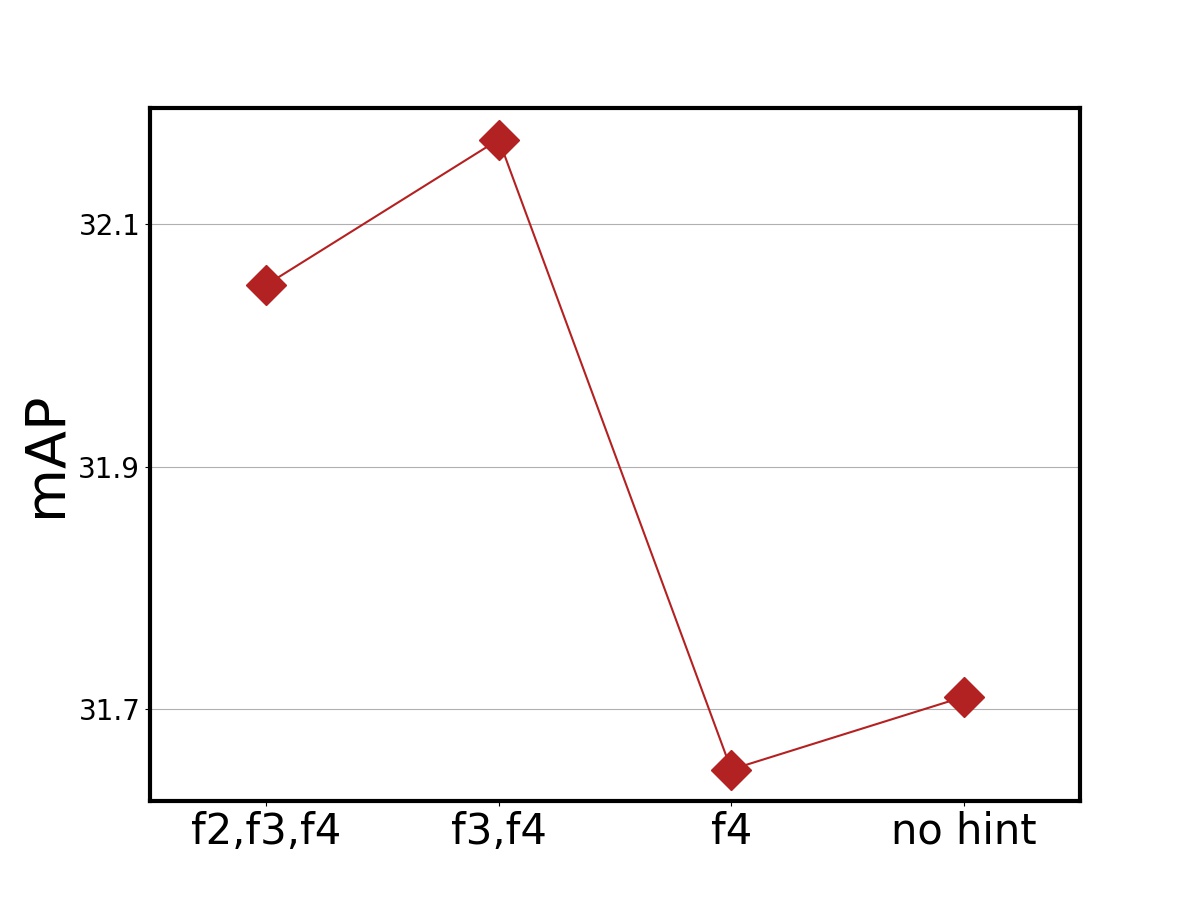}}
\caption{{\bf The ablation study of different feature map hints}. We use the detection transfer result at pruning round 8 (sparsity 83.22\%) as the selection criterion. We observe that using $f_3$ and $f_4$ as feature map hints can induce the best result on detection datasets.
}
\label{fig:feat hint ablation}
\end{center}

\vskip -0.1in
\end{figure}
In this section, we investigate the influence of feature map hints to the detection results. As Figure \ref{fig:feat hint ablation} shows, different feature map hints lead to a different result. We attribute this phenomenon to the different feature maps' detailed and semantic information levels. 
%As the feature map goes higher, it contains much semantic information but has fewer image details. 
If we only use a high-level feature for autoencoder, it takes a lot of network capacity and needs more epochs to converge; if we use a low-level feature map, those features are so close to the original information that the decoder part does not get useful information. We find that using $f_3,f_4$ as feature map hints can lead to the best performance compared to other settings. So we use $f_3$, $f_4$ as the feature map hints choice on other settings.

% \subsection{Longer training epoches and more pruning round can help find universal tickets}
% In Section \ref{sec:autoencoder training}, we introduce feature map hint technique to accelerate the autoencoder training process. We should notice that such a method is focus on training accelerating. If we use longer training epoch and smaller pruning interval

% The key insight of feature map hint: accelerate the training progress. 

\section{Conclusion}
In this paper, we propose a new framework to find the universal tickets that can transfer to diverse downstream datasets. First, we theoretically show that the image-level tasks may result in a degenerated feature map with high probability. This analysis implies that image-level tasks are not sufficient for neural network pruning. To address the problem, we introduce a new pruning framework that includes the image reconstruction task to guide the pruning process. Our framework has three steps: $i$) Train an autoencoder, $ii$) Prune the encoder with an improved LTH method. $iii$) Transfer the encoder to downstream tasks. Besides, the feature map hint method is developed to accelerate the autoencoder training stage. The obtained tickets are evaluated on diverse downstream tasks at different sparsity ratios. The empirical study demonstrates that the proposed method can outperform state-of-the-art method~\cite{chen2021lottery,girish2021lottery} on benchmark datasets.

\newpage
\appendix
\onecolumn

% You can have as much text here as you want. The main body must be at most $8$ pages long.
% For the final version, one more page can be added.
% If you want, you can use an appendix like this one, even using the one-column format.
% %%%%%%%%%%%%%%%%%%%%%%%%%%%%%%%%%%%%%%%%%%%%%%%%%%%%%%%%%%%%%%%%%%%%%%%%%%%%%%%
% %%%%%%%%%%%%%%%%%%%%%%%%%%%%%%%%%%%%%%%%%%%%%%%%%%%%%%%%%%%%%%%%%%%%%%%%%%%%%%%

\section{Proof of Theorem \ref{th:prune LCNN}}
\label{appendix:LCNN}

We use Discrete Fourier Transformation to proof Theorem \ref{th:prune LCNN}

 {\bf Discrete Fourier Transformation} (DFT) of a vector sequence $x \in \mathbb{R}^{c\times N}$is defined as:
 
  \begin{equation}
 \label{eq:fourier}
     \mathcal{F}(x_{i,:})(k) = \tilde{x}_{i}^k = \frac{1}{N}\sum_{s}x_{i,s}e^{\frac{2\pi J}{N}sk}
 \end{equation}
   We use $J$ represent imaginary unit to distinguish from footnote $i$. Here we assume $x_{i,:}$ follow the Periodic Boundary Conditions. 
  DFT has two important propositions:

\begin{proposition} Let $\tilde{x}_{i}^k = \mathcal{F}(x_{i,:})(k)$, then we have:
\\(1) $x_{i,s}= \sum_{k}\tilde{x}_{i}^k e^{\frac{2\pi J}{N}sk}$ \\
(2) $[W*x]_{i,j} = \sum_{t,k}\tilde{W}(k)_{it}\tilde{x}^k_te^{\frac{2\pi J}{ N}jk}$ %\sum_{t}\sum_{k}(\sum_s W_{it,s}e^{\frac{J}{2\pi}sk})\tilde{x}^k_te^{\frac{J}{2\pi}jk} = 
\end{proposition}
We define $[\tilde{W}(k)]_{ij} = \sum_s{W}_{ ij,s}e^{\frac{2\pi J}{ N}sk}$ for simplicity. By those propositions,  $\mathcal{C}(x)$ can be written as a simple version:
\begin{equation}
\label{eq:fourier LCNN}
\mathcal{C}(x)_{:,p} = \sum_{k} W^L(k)W^{L-1}(k)...W^{0}(k)\tilde{x}^k_: e^{\frac{2\pi J}{ N}pk}
\end{equation}
Because the avgpool operation only associate with the $k=0$ component, while other part is independent with it. It's easy to find another network such the zero component unchanged will other component are different. Thus we can proof the claim we stated in Section \ref{sc:Pruning in LCNN}

{\bf claim} Let $\mathcal{C}$ be a LCNN, there must exists another LCNN $C'$, such that
\begin{equation}
    P_{a}\mathcal{C}(x)= P_{a}\mathcal{C'}(x), \mathcal{C}(x)\neq\mathcal{C'}(x)
\end{equation}

To prove Theorem \ref{th:prune LCNN} we should notice that in such a LCNN $\mathcal{C}$, we can find some weights (around $\mathcal{O}(\frac{\epsilon}{\Delta})$) such that $|\sum_sW_{ij,s}| < \epsilon$. By the assumption that each weights is initialized independent, the high frequency component of $W$ is in $\mathcal{O}(1)$ while the zero frequency component is $\mathcal{O}(\epsilon)$.  Therefore, the total influence of $||P_a(\mathcal{C}(x))||^2$ is $\mathcal{O}(\frac{\epsilon^3}{\Delta^3})$ but the influence of $||\mathcal{C}(x)||^2$ is $\mathcal{O}(\frac{\epsilon}{\Delta})$

\begin{lemma}
For a randomly initialized convolutional tensor, $W_{ij,s}\sim i.i.d \mathcal{N}(0,\Delta)$, we have:

(1) $\mathbb{E}(\sum_i\sum_{k=-s}^{s}W_{ij,k}\sum_{n=-s}^{s}W_{im,n}) = (2s+1)n \Delta^2\delta_{jm}$\\
(2) $p = \mathbb{P}(\sum_{k=-s}^{s}W_{ij,k}<\epsilon) < \sqrt{\frac{1}{2\pi(2s+1)}} \frac{\epsilon}{\Delta}$\\

 If we prune every neuron with $|\sum_s W_{ij,s}|<\epsilon$ and let the new tensor is $ W'_{ij,s}$, we have $W'_{ij,s} = m_{ij} W_{ij,s}$\\
 (3) $\mathbb{P}(m_{ij}=0)=p$\\
 (4) $\mathbb{E}(\sum_{t}m_{ij}W_{ij,t}\sum_kW_{ij,k}) > \Delta^2(1-\frac{2\epsilon^3}{3\sqrt{2\pi}\Delta^3})$\\
 (5) $\mathbb{E}(\sum_{t}m_{ij}W_{ij,t}\sum_km_{ij}W_{ij,k}) =\mathbb{E}(\sum_{t}m_{ij}W_{ij,t},\sum_kW_{ij,k})$\\
  (6) $\mathbb{E}(\sum_{t}m_{ij}W_{ij,t}\sin tk\theta\sum_mW_{ij,m}\sin tk\theta) = (1-p)\sum_{t} \sin^2 tk\theta  $\\
 (7) $\mathbb{E}(\sum_{t}m_{ij}W_{ij,t}\cos tk\theta\sum_mW_{ij,m}\cos tk\theta) < (1-p) \sum_{t} \cos^2 tk\theta $\\
in (6)(7) $\theta = \frac{2\pi}{D}$, and $k\in \{1,2,...,D-1\} $
\end{lemma}
(1)(2)(3)(5) are easy to understand, for(4), we have:
\begin{align}
    &\mathbb{E}(\sum_{t}m_{ij}W_{ij,t},\sum_kW_{ij,k}) \\
    =& \mathbb{E}(\sum_t W_{ij,t}\sum_k W_{ij,k}\big | m_{ij}=1) (1-p) + \mathbb{E}(m_{}\sum_t W_{ij,t}\sum_k W_{ij,k}\big | m_{ij}=0) p\\
    =&\Delta^2 - \int_{-\epsilon}^{\epsilon}\frac{1}{\sqrt{2\pi}\Delta} x^{2}e^{-\frac{x^2}{2\pi \Delta}} dx > \Delta^2(1-\frac{2\epsilon^3}{3\sqrt{2\pi}\Delta^3})
\end{align}

for (6) this result is a direcly result by considering the conditional expectation:
\begin{equation}
\mathbb{E}(\sum_{t}W_{ij,t}\sin tk\theta\sum_kW_{ij,k}\sin tk\theta\big ||\sum_tW_{ij,t}|>\epsilon)
\end{equation}
Due to $\sum_{t} 1\cdot\sin tk\theta = 0 $, this expectation is independent of conditional variable. Thus we have:
\begin{align}
    &\mathbb{E}(\sum_{t}m_{ij}W_{ij,t}\sin tk\theta\sum_tW_{ij,t}\sin tk\theta) \\
    =& \mathbb{E}(\sum_{t}m_{ij}W_{ij,t}\sin tk\theta\sum_kW_{ij,k}\sin tk\theta\big | m_{ij}=1) (1-p) \\
    &+ \mathbb{E}(\sum_{t}m_{ij}W_{ij,t}\sin tk\theta\sum_tW_{ij,t}\sin tk\theta\big | m_{ij}=0) p\\
    =& (1-p) \sum_{t} (\sin tk\theta)^2     
\end{align}
For (7), although the expectation term is dependent on the conditional term, we knew that this condition will enlarge the expectation, so we have:
\begin{align}
    &\mathbb{E}(\sum_{t}m_{ij}W_{ij,t}\cos tk\theta\sum_tW_{ij,t}\cos tk\theta)\\
    =& \mathbb{E}(\sum_{t}m_{ij}W_{ij,t}\cos tk\theta\sum_kW_{ij,k}\cos tk\theta\big | m_{ij}=1) (1-p) \\
    &+ \mathbb{E}(\sum_{t}m_{ij}W_{ij,t}\cos tk\theta\sum_tW_{ij,t}\sin tk\theta\big | m_{ij}=0) p\\
    <& (1-p) \sum_{t} (\cos tk\theta )^2 
\end{align}

Now, we can calculate the difference caused by prune the weights:
\begin{lemma}
\label{lem:feature vector}
% \begin{align}
% $\mathbb{E}(\sum_{t}m_{ij}W_{ij,t}\cos tk\theta\sum_mW_{ij,m}\cos tk\theta) > \Delta^2(1-\frac{2\epsilon^3}{3\sqrt{2\pi}\Delta^3})$
% \end{align}

For any random initialized LCNN, where parameter is i.i.d initialized as $\mathcal{N}(0,\Delta)$. If we prune every neuron with $|\sum_s W^l_{ij,s}|<\epsilon$  and get a pruned LCNN $\mathcal{C}'$, then we have \[\frac{||P_a\mathcal{C}(x)-P_a\mathcal{C}(x)||^2}{||P_a\mathcal{C}(x)||^2}\sim\mathcal{O}(\frac{\epsilon^3}{\Delta^3})\] 

\end{lemma}
{\bf proof} According to (\ref{eq:fourier LCNN}), we can rewrite $P_a\mathcal{C}(x)$ as:
\begin{equation}
    P_a\mathcal{C}= W^L(0)W^{L-1}(0)...W^{0}(0)P_a(x)
\end{equation}for simplicity, we write $F^l(k)=W^l(k)W^{l-1}(k)...W^{0}(k)\mathcal{F}(x)(k)$. Thus, 
\begin{equation}
||P_a\mathcal{C}(x)-P_a\mathcal{C}'(x)||^2 = ||F^L(0)-{F'}^L(0)||^2
\end{equation}
We define $\mathbb{E}_l$ as the expectation about the $l$-th layer weights, then we have:

\begin{align}
&\mathbb{E}||P_a\mathcal{C}(x)-P_a\mathcal{C}'(x)||^2\\ = &\mathbb{E}\mathbb{E}_L||F^L(0)-{F'}^L(0)||^2\\
= &\mathbb{E}\mathbb{E}_L\left <F^L(0),{F}^L(0)\right > + 
\mathbb{E}\mathbb{E}_L\left <{F'}^L(0),{F'}^L(0)\right > -
2\mathbb{E}\mathbb{E}_L\left <F^L(0),{F'}^L(0)\right >\\
= &\mathbb{E}\mathbb{E}_L\left<W^L(0)F^{L-1}(0),W^L(0){F}^{L-1}(0)\right > + \mathbb{E}\mathbb{E}_L\left <{W'}^L(0){F'}^{L-1}(0),{W'}^L(0){F'}^{L-1}(0)\right > - \\
&2\mathbb{E}\mathbb{E}_L\left <W^L(0)F^{L-1}(0),{W'}^L(0){F'}^{L-1}(0)\right >\\
< &m_L(2s+1)\Delta^2\left<F^{L-1}(0),{F}^{L-1}(0)\right> + m_L(2s+1)\Delta^2(1-\frac{2\epsilon^3}{3\sqrt{2\pi}\Delta^3})\left <{F'}^{L-1}(0),{F'}^{L-1}(0)\right > - \\
&2m_L(2s+1)\Delta^2(1-\frac{2\epsilon^3}{3\sqrt{2\pi}\Delta^3})\left <F^{L-1}(0),{F'}^{L-1}(0)\right >\\
<& m_Lm_{L-1}...m_{1}(2s+1)^{L}\Delta^{2L}[1-(1-\frac{2\epsilon^3}{3\sqrt{2\pi}\Delta^3})^{L}]\left <P_a(x),P_a(x)\right > \\
<& m_Lm_{L-1}...m_{1}(2s+1)^{L}\Delta^{2L}L\frac{2\epsilon^3}{3\sqrt{2\pi}\Delta^3}\left <P_a(x),P_a(x)\right > \\
=& L\frac{2\epsilon^3}{3\sqrt{2\pi}\Delta^3}\mathbb||P_a\mathcal{C}(x)||^2
\end{align}

By using the concentretion inequality, we proof Lemma \ref{lem:feature vector}

\begin{lemma}
\label{lem:feature map}
% \begin{align}
% $\mathbb{E}(\sum_{t}m_{ij}W_{ij,t}\cos tk\theta\sum_mW_{ij,m}\cos tk\theta) > \Delta^2(1-\frac{2\epsilon^3}{3\sqrt{2\pi}\Delta^3})$
% \end{align}

For any random initialized LCNN, where parameter is i.i.d initialized as $\mathcal{N}(0,\Delta)$. If we prune every neuron with $|\sum_s W^l_{ij,s}|<\epsilon$  and get a pruned LCNN $\mathcal{C}'$, then we have \[\frac{||\mathcal{C}(x)-\mathcal{C}(x)||^2}{||\mathcal{C}(x)||^2}\sim\mathcal{O}(\frac{\epsilon}{\Delta})\] 

\end{lemma}

{\bf proof} First we have 
\[||\mathcal{C}(x)-\mathcal{C}(x)||^2 = \sum_k||\mathcal{F}(\mathcal{C}(x))(k)-\mathcal{F}(\mathcal{C}'(x))(k)||^2\]
We caculate $k\neq 0$ term, due to $k=0$ is calculated in \ref{lem:feature vector} and it is $\mathcal{O}(\frac{\epsilon}{\Delta})$

\begin{align}
&\sum_{k\neq 0} \mathbb{E}||\mathcal{F}(\mathcal{C}(x))(k)-\mathcal{F}(\mathcal{C}'(x))(k)||^2\\
= &\sum_{k\neq 0}\mathbb{E}\mathbb{E}_L||F^L(k)-{F'}^L(k)||^2\\
= &\sum_{k\neq 0}\mathbb{E}\mathbb{E}_L\left <W^L(k)F^{L-1}(k),W^L(k){F}^{L-1}(k)\right > - \mathbb{E}\mathbb{E}_L\left <W^L(k)F^{L-1}(k),{W'}^L(k){F'}^{L-1}(0)\right >\\
> &\sum_{k\neq 0} \mathbb{E}(2s+1)m_L\Delta^2\mathbb{E}\left <F^{L-1}(k),{F}^{L-1}(k)\right > -\\
&\mathbb{E}m_L\Delta^2\mathbb{E}\left <F^{L-1}(k),{F}^{L-1}(k)\right > (1-p)(\sum_{t} \cos^2 tk\theta + \sum_{t} \sin^2 tk\theta)\\
= &\sum_{k\neq 0} m_Lm_{L-1}....m_{1}(2s+1)^L\Delta^{2L}\mathbb{E}\left <\mathcal{F}(x)(k),\mathcal{F}(x)(k)\right >(1 -(1-p)^L)\\ 
= &(1 -(1-p)^L)\sum_{k\neq 0} \mathbb{E}||\mathcal{F}(\mathcal{C}(x))(k)||^2
\end{align}
Thus we have:
\begin{align}
    &\sum_{k} \mathbb{E}||\mathcal{F}(\mathcal{C}(x))(k)-\mathcal{F}(\mathcal{C}'(x))(k)||^2\\
    =&(1 -(1-p)^L)\sum_{k} \mathbb{E}||\mathcal{F}(\mathcal{C}(x))(k)||^2 =\mathcal{O}(\frac{\epsilon}{\Delta})||\mathcal{C}(x)||^2
\end{align}

By combining Lemma \ref{lem:feature vector} and Lemma \ref{lem:feature map}, we proof Theorem \ref{th:prune LCNN}

\section{Proof of Theorem \ref{th:finetune OR}}
\label{appendix:ORCNN}

In order to prove Theorem \ref{th:finetune OR}, we need more notations to represent the problem setup.

First, we introduce the convolution expansion operator $\phi_{\cdot}(\cdot)$ that can simplify convolution operator to inner product of tensors.

{\bf Convolution Tensor} In convolutional neural network(CNN), the convolution tensor is described as follows. Let ${\mathbf{w}}_{ij}=(w_{ij,-s},w_{ij,-s+1},...,w_{ij,s})^{\mathbf{T}} \in \mathbb{R}^{2s+1}(1 \leq i \leq c'$, $1 \leq j \leq c)$ be convolution kernel, where $c'$ is the number of filters and $2s+1$ is the size of convolution kernel. And $\mathbf{W}_{i}=(\mathbf{w}_{i1},\mathbf{w}_{i2},...,\mathbf{w}_{ic})^{\mathbf{T}} \in \mathbb{R}^{c \times (2s+1)}$ denotes $i_{th}$ filter. Then the convolution tensor is defined as a 3-dimensional tensor $\mathbf{W}=(\mathbf{W}_{i})_{1 \leq i \leq c'} \in \mathbb{R}^{c' \times c \times (2s+1)}$.

{\bf Convolution Expansion Operator}  Let $\mathbf{x} = (x_{i,j}) \in \mathbb{R}^{c \times D}$ be the 1-dimensional input, where $D$ is the length of input sequence and c is the channel number of input. Let $\mathbf{W} \in \mathbb{R}^{c' \times c \times (2s+1)}$ be the convolution tensor. Then we define $\phi_{k}(\mathbf{x})$ as
\begin{equation}
    \begin{aligned}  \phi_{k}\left(\mathbf{x}\right) =&\left(\begin{array}{ccc} {x}_{1,k-s} ,&  \ldots & , {x}_{1,k+s}\\ \ldots, & \ldots, & \ldots \\ {x}_{c,k-s}, & \ldots, & {x}_{c,k+s}\end{array}\right) \end{aligned}
\end{equation}
Recall the definition of convolution operator, we have
\begin{equation}
    (\mathbf{W}*\mathbf{x})_{r,k} = \left<\mathbf{W}_r,\phi_{k}(\mathbf{x})\right>
\end{equation}

Because we use structured pruning method to prune model, we can assume the new channel number of feature maps is $M=m q=m(1-p)$, where $p$ is our pruning rate. And we assume $m q$ is an integer in order to simplify the proof. Now, we can represent the pruned ORCNN as 
\begin{equation}
    f(\mathbf{x})=\frac{\sqrt{q}}{\sqrt{M}D}\sum_{r=1}^{M}a_r\sum_{k=1}^{D}\sigma(\left<\mathbf{W}_r,\phi_{k}(\mathbf{x})\right>)
\end{equation}

To analyze the gradient descent process, inspired by \cite{pmlr-v97-du19c}, we focus on the Gram matrix $\mathbf{G}=(\mathbf{G}_{i,j})$ of ORCNN, which is defined as
\begin{equation}
    \mathbf{G}_{i,j}=\frac{q}{MD^{2}}\sum_{r=1}^{M}\sum_{k=1}^{D}\sum_{l=1}^{D}\left<\phi_{k}(\mathbf{x}_i),\phi_{l}(\mathbf{x}_j)\right>\mathbb{I}\{ \left<\mathbf{W}_r,\phi_{k}(\mathbf{x}_i)\right> \geq 0,
    \left<\mathbf{W}_r,\phi_{l}(\mathbf{x}_j)\right> \geq 0\}
\end{equation}
And if we consider the case that $M$ goes to infinity, we can represent the expectation of $\mathbf{G}$ and we know $\mathbf{G}$ converges to it by central limit theorem and independence of $\mathbf{W}_r(1 \leq r \leq M)$. We use $\mathbf{G}^{\infty}$ to denote the expectation, which can be formally defined as
\begin{equation}
    \mathbf{G}_{i,j}^{\infty}=\mathbb{E}_{\mathbf{W} \sim \mathbf{N}(\mathbf{0},\mathbf{I})} \left[ \frac{q}{D^{2}}\sum_{k=1}^{D}\sum_{l=1}^{D}\left<\phi_{k}(\mathbf{x}_i),\phi_{l}(\mathbf{x}_j)\right>\mathbb{I}\{ \left<\mathbf{W},\phi_{k}(\mathbf{x}_i)\right> \geq 0,
    \left<\mathbf{W},\phi_{l}(\mathbf{x}_j)\right> \geq 0 \} \right]
\end{equation}

\begin{proposition}
We use $\lambda_{min}(\cdot)$ to denote the least eigenvalue of a matrix, then we assume that $\mathbf{G}^{\infty}$ has the positive least eigenvalue $\lambda_0=\lambda_{min}(\mathbf{G}^{\infty})$. In fact, we know $\mathbf{G}^{\infty}$ is positive definite. Thus, the assumption is weak and reasonable.
\end{proposition}

We use $\mathbf{G}_0$ to denote the initial Gram matrix of pruned model. By the standard concentration analysis of independent variables, we can derive the following Lemma \ref{le:G0}, which means the initial Gram matrix of model also has the bounded positive least eigenvalue with high probability.
\begin{lemma}
\label{le:G0}
When the channel number of feature maps in pruned model $M=\Omega\left(\frac{n^{2}}{\lambda_{0}^{2}} \log \left(\frac{n}{\delta}\right)\right)$ and initialization in pre-trained phase satisfies the standard Guassian distribution, with probability at least $1-\delta$, we have 
\begin{equation}
    \lambda_{min}(\mathbf{G}_0) \geq \frac{3}{4}\lambda_0
\end{equation}
\end{lemma}

Before analyzing the finetuning process, we assume some settings as follows to simplify the process in order to focus on the change of convolution tensor.
\begin{proposition}
In finetuning phase, we randomly initialize the fully connected weight $\mathbf{a} \sim Unif(\{-1,1\}^{M})$. And we normalize the input such that $||\phi_k(\mathbf{x}_i)||=1$ for any $k,i$. In fact, without the normalization assumption we can also proof the similar conclusion, but it will dependent with $\frac{\max_{k,i}\{||\phi_k(\mathbf{x}_i)||\}}{\min_{k,i}\{||\phi_k(\mathbf{x}_i)||\}}$, which is also an constant.
\end{proposition}

Because the fully connected weight $\mathbf{a}$ is fixed, we can represent the model output w.r.t. the given dataset $S$ as 
\begin{equation}
    F_i(\mathbf{W}(t))=\frac{\sqrt{q}}{\sqrt{M}D}\sum_{r=1}^{M}a_r\sum_{k=1}^{D}\sigma(\left<\mathbf{W}_r(t),\phi_{k}(\mathbf{x}_i)\right>)
\end{equation}
which is output of $i^{th}$ sample in dataset. And we use $\mathbf{F}(\mathbf{W}(t))=\left(F_1(\mathbf{W}(t)),F_2(\mathbf{W}(t)),\cdots,F_n(\mathbf{W}(t))\right)^{\mathbf{T}}$ to denote the output vector, where $t$ means it is $t^{th}$-round.

\begin{proposition}
\label{pr:W_0}
Recall the initialization of the pruned convolution tensor is the same as that in pre-trained phase, we know each unpruned entry of the pruned convolution tensor satisfies indenpendent standard Guassian distribution $\mathbf{N}(0,1)$. With the knowledge of sub-exponential distribution, we can prove that w.h.p. the $l_2$ norm of the pruned convolution tensor$\mathbf{W}(0)$ is close to $\sqrt{q m D(2s+1)}$ and the $l_2$ norm of the original convolution tensor before pre-trained phase is close to $\sqrt{m D(2s+1)}$.
\end{proposition}

Next, we introduce Lemma \ref{le:main} that describes the dynamic process in finetuning phase.

\begin{lemma}
\label{le:main}
Under the above assumptions, if the channel number of feature maps in pruned model $M$ is $\Omega\left(\frac{n^{6}}{\lambda_{0}^{4} \delta^{3}}\right)$ and learning rate $\eta$ is $O\left(\frac{\lambda_{0}}{n^{2}}\right)$, with probability at least $1-\delta$, we have
\begin{equation}
    ||\mathbf{F}(\mathbf{W}(t))-\mathbf{y}||^{2} \leq \left(1-\frac{\eta\lambda_0}{2}\right)^{t} ||\mathbf{F}(\mathbf{W}(0))-\mathbf{y}||^{2} 
\end{equation}
where $\mathbf{y}=(y_1,y_2,\cdots,y_n)^{\mathbf{T}}$ is the label vector of dataset $S$.
\end{lemma}

By Lemma \ref{le:main}, with the standard analysis of gradient descent, we can prove the following Lemma \ref{le:W_fin}
\begin{lemma}
\label{le:W_fin}
Under the above assumptions, if the channel number of feature maps in pruned model $M$ is $\Omega\left(\frac{n^{6}}{\lambda_{0}^{4} \delta^{3}}\right)$ and learning rate $\eta$ is $O\left(\frac{\lambda_{0}}{n^{2}}\right)$, with probability at least $1-\delta$, for some constant $C$, we have
\begin{equation}
    ||\mathbf{W}_{fin}-\mathbf{W}(0)||
    \leq
    \frac{C \sqrt{q} n}{\lambda_0}
\end{equation}
\end{lemma}

{\bf Remark of Lemmma \ref{le:W_fin}} In fact, we have similar conclusion of Lemmma \ref{le:W_fin} in pre-trained phase, which implies $||\mathbf{W}_{pre}||=\sqrt{m}(1+o(1))$ and can be proven by the same method as follows.

{\bf Proof of Theorem \ref{th:finetune OR}} By $m=\Omega(\frac{n^{6}}{\lambda_{0}^{4}})$, Proposition \ref{pr:W_0} and Lemma \ref{le:W_fin}, with probability $1-\delta$, we have $||\mathbf{W}_{fin}-\mathbf{W}(0)||=o(||\mathbf{W}(0)||)$ and $||\mathbf{W}_{fin}||=\sqrt{q m}(1+o(1))$. Combined with the property of rotation operator, for any rotation operator $\mathbf{Q}$, we have $ dist_{l_2}^{N}(\mathbf{Q}\mathbf{W}_{fin},\mathbf{W}_{pre}) \geq
    \frac{||\mathbf{W}_{pre}||-||\mathbf{Q}\mathbf{W}_{fin}||}{\sqrt{||\mathbf{W}_{pre}||||\mathbf{Q}\mathbf{W}_{fin}||}} = (1-p)^{-\frac{1}{4}}-(1-p)^{\frac{1}{4}} \geq \frac{p}{2}$.
Thus, we prove Theorem \ref{th:finetune OR} .
%\begin{equation}
%    dist_{l_2}^{N}(\mathbf{Q}\mathbf{W}_{fin},\mathbf{W}_{pre}) \geq
%    \frac{||\mathbf{W}_{pre}||-||\mathbf{Q}\mathbf{W}_{fin}||}{\sqrt{||\mathbf{W}_{pre}||||\mathbf{Q}\mathbf{W}_{fin}||}} = (1-p)^{-\frac{1}{4}}-(1-p)^{\frac{1}{4}} \geq \frac{p}{2}
%\end{equation}

%By Lemma \ref{le:main}, with the standard analysis of gradient descent, we can prove w.h.p. $||\mathbf{W}_{fin}-\mathbf{W}_0||$ is $O(1)$ such that $||\mathbf{W}_{fin}||$ is also close to $\sqrt{q m D(2s+1)}$. And we can derive w.h.p $||\mathbf{W}_{pre}||$ is close to $\sqrt{m D(2s+1)}$. By the property of rotation operator, for
%any rotation operator $\mathbf{Q}$, $||\mathbf{Q}\mathbf{W}_{fin}||$ is also close to $\sqrt{q m D(2s+1)}$. Thus, under the over-parameterized assumption, with high probability, we have $\min_{\mathbf{Q} \in \mathbb{O}} \{dist_{l_2}^{N}(\mathbf{Q}\mathbf{W}_{fin},\mathbf{W}_{pre})\} \geq (1-p)^{-\frac{1}{4}}-(1-p)^{\frac{1}{4}} \approx \frac{p}{2}$.

Next, We will use induction method to prove Lemma \ref{le:main} and Lemma \ref{le:W_fin} . Our inductive hypothesis is the inequality is true for $0,1,\cdots,t$ and we want to prove the inequality is also true for $t+1$. First, we have the following decomposition of $l_2$ loss. 
\begin{equation}
\begin{aligned}
    &||\mathbf{F}(\mathbf{W}(t+1))-\mathbf{y}||^{2}\\ 
    &= ||\mathbf{F}(\mathbf{W}(t+1))-\mathbf{F}(\mathbf{W}(t))+\mathbf{F}(\mathbf{W}(t))-\mathbf{y}||^{2}
    \\
    &= ||\mathbf{F}(\mathbf{W}(t+1))-\mathbf{F}(\mathbf{W}(t))||^{2}
    \\
    &\quad +2\left(\mathbf{F}(\mathbf{W}(t))-\mathbf{y}\right)^{\mathbf{T}}\left(\mathbf{F}(\mathbf{W}(t+1))-\mathbf{F}(\mathbf{W}(t))\right)
    +||\mathbf{F}(\mathbf{W}(t))-\mathbf{y}||^{2}
\end{aligned}
\end{equation}
The third term can be bounded by the inductive hypothesis. Thus, we only need to bound the first and the second terms.

The following Lemma \ref{le:grad} can help us to bound the gradient norm in finetuing phase.
\begin{lemma}
\label{le:grad}
In finetuing phase, we can control the upper bound of gradient norm as
\begin{equation}
    \left|\left|\frac{\partial L(\mathbf{W})}{\partial \mathbf{W}_r}\right|\right|
    \leq \frac{\sqrt{q n}}{\sqrt{M}}||\mathbf{F}(\mathbf{W})-\mathbf{y}||
\end{equation}
for any $r \in [M]$.
\end{lemma}
{\bf Proof of Lemma \ref{le:grad}} The proof is very standard to analysis gradient descent.
\begin{equation}
\begin{aligned}
    \left|\left|\frac{\partial L(\mathbf{W})}{\partial \mathbf{W}_r}\right|\right|
    &= \left|\left|\frac{\sqrt{q} }{\sqrt{M}D}\sum_{i=1}^{n}(F_i(\mathbf{W})-y_i)\sum_{k=1}^{D}\phi_{k}(\mathbf{x}_i)\mathbb{I}\{ \left<\mathbf{W}_r,\phi_{k}(\mathbf{x}_i)\right> \geq 0 \} \right|\right|
    \\
    & \leq \frac{\sqrt{q} }{\sqrt{M}D}\left|\left|\mathbf{F}(\mathbf{W})-\mathbf{y}\right|\right|
    \left|\left|\left(\sum_{k=1}^{D}\phi_{k}(\mathbf{x}_i)\mathbb{I}\{ \left<\mathbf{W}_r,\phi_{k}(\mathbf{x}_i)\right> \geq 0 \} \right)_{1 \leq i \leq n}\right|\right|
    \\
    & = \frac{\sqrt{q} }{\sqrt{M}D}||\mathbf{F}(\mathbf{W})-\mathbf{y}||
    \sqrt{\sum_{i=1}^{n}||\sum_{k=1}^{D}\phi_{k}(\mathbf{x}_i)\mathbb{I}\{ \left<\mathbf{W}_r,\phi_{k}(\mathbf{x}_i)\right> \geq 0 \} ||^{2}}
    \\
    & \leq \frac{\sqrt{q} }{\sqrt{M}D}||\mathbf{F}(\mathbf{W})-\mathbf{y}||
    \sqrt{\sum_{i=1}^{n}\left(\sum_{k=1}^{D}||\phi_{k}(\mathbf{x}_i)\mathbb{I}\{ \left<\mathbf{W}_r,\phi_{k}(\mathbf{x}_i)\right> \geq 0 \}|| \right)^{2}}
    \\
    & \leq \frac{\sqrt{q} }{\sqrt{M}D}||\mathbf{F}(\mathbf{W})-\mathbf{y}||
    \sqrt{n D^{2}}
    \\
    & = \frac{\sqrt{q n}}{\sqrt{M}}||\mathbf{F}(\mathbf{W})-\mathbf{y}||
\end{aligned}
\end{equation}

Now, we can derive an upper bound of the first term by Lemma \ref{le:grad} and 1-Lipschitz property of ReLU funciton.
\begin{equation}
\begin{aligned}
     &||\mathbf{F}(\mathbf{W}(t+1))-\mathbf{F}(\mathbf{W}(t))||^{2}\\
    &=
    \sum_{i=1}^{n}\left|F_i(\mathbf{W}(t+1))-F_i(\mathbf{W}(t))\right|^{2}
    \\
    &=
    \sum_{i=1}^{n}\left| \frac{\sqrt{q}}{\sqrt{M}D}\sum_{r=1}^{M}a_r\sum_{k=1}^{D}\left( \sigma(\left<\mathbf{W}_r(t+1),\phi_{k}(\mathbf{x}_i)\right>)-\sigma(\left<\mathbf{W}_r(t),\phi_{k}(\mathbf{x}_i)\right>)\right) \right|^{2}
    \\
    &=
     \sum_{i=1}^{n}\left| \frac{\sqrt{q}}{\sqrt{M}D}\sum_{r=1}^{M}a_r\sum_{k=1}^{D}\left( \sigma\left(\left<\mathbf{W}_r(t)-\eta \frac{\partial L(\mathbf{W(t)})}{\partial \mathbf{W(t)}_r},\phi_{k}(\mathbf{x}_i)\right>\right)-\sigma(\left<\mathbf{W}_r(t),\phi_{k}(\mathbf{x}_i)\right>)\right) \right|^{2}
     \\
     & \leq
     \frac{q}{D}
     \sum_{i=1}^{n} \sum_{r=1}^{M} \sum_{k=1}^{D}\left|
     \left<\eta \frac{\partial L(\mathbf{W}(t))}{\partial \mathbf{W}_r(t)},\phi_{k}(\mathbf{x}_i)\right>
     \right|^{2}
     \\
     & \leq
     \frac{q{\eta}^{2}}{D}
     \sum_{i=1}^{n} \sum_{r=1}^{M} \sum_{k=1}^{D}
     \left|\left|\frac{\partial L(\mathbf{W}(t))}{\partial \mathbf{W}_r(t)}\right|\right|^{2}
     \\
     & \leq
     \frac{q{\eta}^{2}}{D} n M D \frac{q n}{M}||\mathbf{F}(\mathbf{W}(t))-\mathbf{y}||^{2}
     \\
     & =
     q^{2} {\eta}^{2} n^{2} ||\mathbf{F}(\mathbf{W}(t))-\mathbf{y}||^{2}
\end{aligned}
\end{equation}

To bound the second term, we need a more refined analysis of its form. The following Lemma \ref{le: w_r(t)} will inspire us how to deal with the term.
\begin{lemma}
\label{le: w_r(t)}
Under the inductive hypothesis, for any $T \in [t+1]$ and $r in [M]$, we have
\begin{equation}
    ||\mathbf{W}_r(T)-\mathbf{W}_r(0)|| \leq
    \frac{4\sqrt{q n}}{\sqrt{M} \lambda_0}
    ||\mathbf{F}(\mathbf{W}(0))-\mathbf{y}||
\end{equation}
\end{lemma}

{\bf Proof of Lemma \ref{le: w_r(t)}} By the gradient descent and inductive hypothesis, we have
\begin{equation}
\begin{aligned}
    ||\mathbf{W}_r(T)-\mathbf{W}_r(0)||
    & \leq
    \sum_{j=0}^{T-1}||\mathbf{W}_r(j+1)-\mathbf{W}_r(j)||
    \\
    & =
    {\eta}\sum_{j=0}^{T-1} \left|\left|\frac{\partial L(\mathbf{W}(j))}{\partial \mathbf{W}_r(j)}\right|\right|
    \\
    & \leq
    {\eta}\sum_{j=0}^{T-1} \frac{\sqrt{q n}}{\sqrt{M}}||\mathbf{F}(\mathbf{W}(j))-\mathbf{y}||
    \\
    & \leq
    {\eta}\sum_{j=0}^{T-1} \frac{\sqrt{q n}}{\sqrt{M}}\left(1-\frac{\eta\lambda_0}{2}\right)^{\frac{j}{2}}||\mathbf{F}(\mathbf{W}(0))-\mathbf{y}||
    \\
    & \leq
    {\eta}\frac{\sqrt{q n}}{\sqrt{M}}||\mathbf{F}(\mathbf{W}(0))-\mathbf{y}||
    \sum_{j=0}^{\infty}\left(1-\frac{\eta\lambda_0}{2}\right)^{\frac{j}{2}}
    \\
    & \leq
    {\eta}\frac{\sqrt{q n}}{\sqrt{M}}||\mathbf{F}(\mathbf{W}(0))-\mathbf{y}||
    \sum_{j=0}^{\infty}\left(1-\frac{\eta\lambda_0}{4}\right)^{j}
    \\
    & \leq
    \frac{4\sqrt{q n}}{\sqrt{M} \lambda_0}
    ||\mathbf{F}(\mathbf{W}(0))-\mathbf{y}||
\end{aligned}
\end{equation}

Notice we have
\begin{equation}
    \begin{aligned}
        F_i(\mathbf{W}(t+1))-F_i(\mathbf{W}(t))
        &= \frac{\sqrt{q}}{\sqrt{M}D}\sum_{r=1}^{M}a_r\sum_{k=1}^{D}\left( \sigma(\left<\mathbf{W}_r(t+1),\phi_{k}(\mathbf{x}_i)\right>)-\sigma(\left<\mathbf{W}_r(t),\phi_{k}(\mathbf{x}_i)\right>)\right)
    \end{aligned}
\end{equation}

And Lemma \ref{le: w_r(t)} implies the distance between $\mathbf{W}_r(T)$ and $\mathbf{W}_r(0)$ is bounded, we can use truncation estimation method to deal with $F_i(\mathbf{W}(t+1))-F_i(\mathbf{W}(t))$. Specifically, we use $U_{r,k,i}(R)$ to denote the event $\left<\mathbf{W}_r(0),\phi_{k}(\mathbf{x}_i)\right> < R$.

Because $\phi_{k}(\mathbf{x}_i)$ is normalized, we know $\left<\mathbf{W}_r(0),\phi_{k}(\mathbf{x}_i)\right>$ is a standard Guassian Variable. We have the following Lemma \ref{le:U} .

\begin{lemma}
\label{le:U}
For any $r \ in [M], k \in [D]$ and $i \in [n]$, we have the probability of the $U_{r,k,i}(R)$ satisfies
\begin{equation}
    \mathbb{P}\{U_{r,k,i}(R)\} \leq \frac{2R}{\sqrt{2 \pi}}
\end{equation}
\end{lemma}

If the event $U_{r,k,i}(R)$ is false and $\frac{4\sqrt{q n}}{\sqrt{M} \lambda_0}
    ||\mathbf{F}(\mathbf{W}(0))-\mathbf{y}|| < R$, then we know $\mathbb{I}\{ \left<\mathbf{W}_r(t+1),\phi_{k}(\mathbf{x}_i)\right> \geq 0 \}=\mathbb{I}\{ \left<\mathbf{W}_r(t),\phi_{k}(\mathbf{x}_i)\right> \geq 0 \}=\mathbb{I}\{ \left<\mathbf{W}_r(0),\phi_{k}(\mathbf{x}_i)\right> \geq 0 \}$ by Lemma \ref{le: w_r(t)} . So we use $J_{true}(R)$ to denote the set $\{(r,k,i)|U_{r,k,i}(R)$ is true$\}$ and $J_{false}(R)$ to denote the set $\{(r,k,i)|U_{r,k,i}(R)$ is false$\}$.

By Markov's inequality, we can derive the following Lemma \ref{le:J}
\begin{lemma}
\label{le:J}
With probability at least $1-\delta$, we have
\begin{equation}
    |J_{true}(R)| \leq \frac{n M D R}{\sqrt{2 \pi} \delta}
\end{equation}
\end{lemma}

If we use $\mathbf{e}_i(1 \leq i \leq n)$ to denote the standard basis of $\mathbb{R}^{n}$, then we can decompose $\mathbf{F}(\mathbf{W}(t+1))-\mathbf{F}(\mathbf{W}(t))$ as
\begin{equation}
\begin{aligned}
    &\mathbf{F}(\mathbf{W}(t+1))-\mathbf{F}(\mathbf{W}(t))\\
    &=
    \sum_{i=1}^{n}\mathbf{e}_i \frac{\sqrt{q}}{\sqrt{M}D}\sum_{r=1}^{M}a_r\sum_{k=1}^{D}\left( \sigma(\left<\mathbf{W}_r(t+1),\phi_{k}(\mathbf{x}_i)\right>)-\sigma(\left<\mathbf{W}_r(t),\phi_{k}(\mathbf{x}_i)\right>)\right)
    \\
    &= \frac{\sqrt{q}}{\sqrt{M}D}\sum_{(r,k,i) \in J_{true}(R)}a_r\mathbf{e}_i\left( \sigma(\left<\mathbf{W}_r(t+1),\phi_{k}(\mathbf{x}_i)\right>)-\sigma(\left<\mathbf{W}_r(t),\phi_{k}(\mathbf{x}_i)\right>)\right)
    \\
    & \quad +
    \frac{\sqrt{q}}{\sqrt{M}D}\sum_{(r,k,i) \in J_{false}(R)}a_r\mathbf{e}_i\left( \sigma(\left<\mathbf{W}_r(t+1),\phi_{k}(\mathbf{x}_i)\right>)-\sigma(\left<\mathbf{W}_r(t),\phi_{k}(\mathbf{x}_i)\right>)\right)
\end{aligned}
\end{equation}
We use $I_1,I_2$ to denote two terms indexed by $J_{true}(R),J_{false}(R)$. And the term indexed by $J_{true}(R)$ can be bounded by the cardinal of $J_{true}(R)$ as
\begin{equation}
\begin{aligned}
    ||I_1|| &= \left|\left| \frac{\sqrt{q}}{\sqrt{M}D}\sum_{(r,k,i) \in J_{true}(R)}a_r\mathbf{e}_i\left( \sigma(\left<\mathbf{W}_r(t+1),\phi_{k}(\mathbf{x}_i)\right>)-\sigma(\left<\mathbf{W}_r(t),\phi_{k}(\mathbf{x}_i)\right>)\right) \right|\right|
    \\
    &=
    \frac{\sqrt{q}}{\sqrt{M}D}
    \sqrt{\sum_{i=1}^{n}\left|\sum_{r,k|
        (r,k,i) \in J_{true}(R)
    }a_r\left( \sigma(\left<\mathbf{W}_r(t+1),\phi_{k}(\mathbf{x}_i)\right>)-\sigma(\left<\mathbf{W}_r(t),\phi_{k}(\mathbf{x}_i)\right>)\right) \right|^{2}}
    \\
    &=
    \frac{\sqrt{q}}{\sqrt{M}D}
    \sqrt{\sum_{i=1}^{n}\left|\sum_{r,k|
        (r,k,i) \in J_{true}(R)
    }a_r\left( \sigma\left(\left<\mathbf{W}_r(t)-\eta \frac{\partial L(\mathbf{W}(t))}{\partial \mathbf{W}_r(t)},\phi_{k}(\mathbf{x}_i)\right>\right)-\sigma(\left<\mathbf{W}_r(t),\phi_{k}(\mathbf{x}_i)\right>)\right) \right|^{2}}
    \\
    & \leq
    \frac{\sqrt{q}}{\sqrt{M}D}
    \sqrt{\sum_{i=1}^{n}\left(\sum_{r,k|
        (r,k,i) \in J_{true}(R)
    }\eta \left|\left|\frac{\partial L(\mathbf{W}(t))}{\partial \mathbf{W}_r(t)}\right|\right| \right)^{2}}
    \\
    & \leq
    \frac{\sqrt{q} |J_{true}(R)|}{\sqrt{M}D}\eta \left|\left|\frac{\partial L(\mathbf{W}(t))}{\partial \mathbf{W}_r(t)}\right|\right|
    \\
    & \leq
    \frac{\sqrt{q} |J_{true}(R)|}{\sqrt{M}D}\eta
    \frac{\sqrt{q n}}{\sqrt{M}}||\mathbf{F}(\mathbf{W}(t))-\mathbf{y}||
    \\
    & =
    \frac{\eta q \sqrt{n} |J_{true}(R)|}{M D}||\mathbf{F}(\mathbf{W}(t))-\mathbf{y}||
\end{aligned}
\end{equation}

Next, we focus on $I_2$ and establish relation between it and the Gram matrix.
\begin{equation}
\begin{aligned}
    I_2 &=
    \frac{\sqrt{q}}{\sqrt{M}D}\sum_{(r,k,i) \in J_{false}(R)}a_r\mathbf{e}_i\left( \sigma(\left<\mathbf{W}_r(t+1),\phi_{k}(\mathbf{x}_i)\right>)-\sigma(\left<\mathbf{W}_r(t),\phi_{k}(\mathbf{x}_i)\right>)\right)
    \\ 
    &= \frac{\sqrt{q}}{\sqrt{M}D}\sum_{(r,k,i) \in J_{false}(R)}a_r\mathbf{e}_i\left( \sigma\left(\left<\mathbf{W}_r(t)-\eta \frac{\partial L(\mathbf{W}(t))}{\partial \mathbf{W}_r(t)},\phi_{k}(\mathbf{x}_i)\right>\right)-\sigma(\left<\mathbf{W}_r(t),\phi_{k}(\mathbf{x}_i)\right>)\right)
    \\
    &=
    \frac{\sqrt{q}}{\sqrt{M}D}\sum_{(r,k,i) \in J_{false}(R)}a_r\mathbf{e}_i\left( -\eta \frac{\partial L(\mathbf{W}(t))}{\partial \mathbf{W}_r(t)}\right)\mathbb{I}\{ \left<\mathbf{W}_r,\phi_{k}(\mathbf{x}_i)\right> \geq 0\}
    \\
    &=
    -\eta \frac{q}{MD^{2}}\sum_{(r,k,i) \in J_{false}(R)}\sum_{j=1}^{n}\sum_{l=1}^{D}\mathbf{e}_i
    (F_j(\mathbf{W}(t))-y_j)
    \left<\phi_{k}(\mathbf{x}_i),\phi_{l}(\mathbf{x}_j)\right>\mathbb{I}\{ \left<\mathbf{W}_r,\phi_{k}(\mathbf{x}_i)\right> \geq 0,
    \left<\mathbf{W}_r,\phi_{l}(\mathbf{x}_j)\right> \geq 0\}
    \\
    &=
    -\eta \sum_{i=1}^{n}\mathbf{e}_i\sum_{j=1}^{n}\tilde{\mathbf{G}}_{i,j}(t)(F_j(\mathbf{W}(t))-y_j)
    \\
    &=
    -\eta \tilde{\mathbf{G}}(t)(\mathbf{F}(\mathbf{W}(t))-\mathbf{y})
\end{aligned}
\end{equation}
where $\tilde{\mathbf{G}}(t)$ is defined as
\begin{equation}
    \tilde{\mathbf{G}}_{i,j}(t) =
    \frac{q}{MD^{2}}\sum_{r,k,l|(r,k,i) \in J_{false}(R)}\left<\phi_{k}(\mathbf{x}_i),\phi_{l}(\mathbf{x}_j)\right>\mathbb{I}\{ \left<\mathbf{W}_r(t),\phi_{k}(\mathbf{x}_i)\right> \geq 0,
    \left<\mathbf{W}_r(t),\phi_{l}(\mathbf{x}_j)\right> \geq 0\}
\end{equation}

By matrix perturbation technique, we can prove the following Lemma \ref{le:G_t}
\begin{lemma}
\label{le:G_t}
If the channel number $M$ is $\Omega\left(\frac{q^{3}n^{6}}{{\delta}^{2}{\lambda_0}^{4}}\right)$, then with probability at least $1-\delta$, we have
\begin{equation}
    \lambda_{min}(\mathbf{G}(t)) \geq \frac{\lambda_0}{2}
\end{equation}
\end{lemma}

{\bf Proof of Lemma \ref{le:G_t}} First, by $M = \Omega\left(\frac{q^{3}n^{6}}{{\delta}^{2}{\lambda_0}^{4}}\right)$ and Lemma \ref{le: w_r(t)}, if we set $R'$ to $\frac{4\sqrt{q n}}{\sqrt{M} \lambda_0}
    ||\mathbf{F}(\mathbf{W}(0))-\mathbf{y}||$, then we have 
\begin{equation}
    ||\mathbf{W}_r(t)-\mathbf{t}_r(0)||
    \leq
    R'
\end{equation}
for any $r \in [M]$.

Next, we establish relation between $\mathbf{G}(t)$ and $\mathbf{G}(0)$. We consider the difference of each entry of them as
\begin{equation}
\begin{aligned}
     &\mathbb{E}|\mathbf{G}_{i,j}(t)-\mathbf{G}_{i,j}(0)|\\
     &=
     \frac{q}{MD^{2}}\sum_{r=1}^{M}\sum_{k=1}^{D}\sum_{l=1}^{D}|\left<\phi_{k}(\mathbf{x}_i),\phi_{l}(\mathbf{x}_j)\right>(\mathbb{I}\{ \left<\mathbf{W}_r(t),\phi_{k}(\mathbf{x}_i)\right> \geq 0,
    \left<\mathbf{W}_r(t),\phi_{l}(\mathbf{x}_j)\right> \geq 0\}
    \\
    & \quad
    -\mathbb{I}\{ \left<\mathbf{W}_r(0),\phi_{k}(\mathbf{x}_i)\right> \geq 0, \left<\mathbf{W}_r(0),\phi_{l}(\mathbf{x}_j)\right> \geq 0 \} )|
    \\
    & \leq
    \frac{q}{MD^{2}}\sum_{r=1}^{M}\sum_{k=1}^{D}\sum_{l=1}^{D} \mathbb{P}\{U_{r,k,i}(R') \cup U_{r,L,i}(R') \}
    \\
    & \leq
    \frac{4q R'}{\sqrt{2 \pi}}
\end{aligned}
\end{equation}
By the Markov's inequality, with probability at least $1-\delta$, we have $||\mathbf{G}(t)-\mathbf{G}(0)||_{l_{1}} \leq \frac{4q n^{2} R'}{\sqrt{2 \pi} \delta}$. And by the positive definite of $\mathbf{G}(t),\mathbf{G}(0)$we know
\begin{equation}
    \lambda_{max}(\mathbf{G}(t)-\mathbf{G}(0)) \leq ||\mathbf{G}(t)-\mathbf{G}(0)||_{F} \leq ||\mathbf{G}(t)-\mathbf{G}(0)||_{l_{1}} \leq \frac{4q n^{2} R'}{\sqrt{2 \pi} \delta}
\end{equation}
Thus, we have
\begin{equation}
\begin{aligned}
    \lambda_{min}(\mathbf{G}(t)) &\geq \lambda_{min}(\mathbf{G}(0)) - \lambda_{max}(\mathbf{G}(t)-\mathbf{G}(0))
    \\
    & \geq
    \frac{3}{4}\lambda_0 - \frac{4q n^{2} R'}{\sqrt{2 \pi} \delta} \geq 
    \frac{\lambda_0}{2}
\end{aligned}
\end{equation}

Under conclusion of Lemma \ref{le:G_t}, by the similar technique used by proof of Lemma \ref{le:G_t} and Lemma \ref{le:J}, we can prove
\begin{lemma}
\label{le:G_c}
If the channel number $M$ is $\Omega\left(\frac{q^{3}n^{6}}{{\delta}^{2}{\lambda_0}^{4}}\right)$, then with probability at least $1-\delta$, we have
\begin{equation}
    \lambda_{min}(\tilde{\mathbf{G}}(t)) \geq \frac{\lambda_0}{2}-\frac{q n^{2} R}{\sqrt{2 \pi} \delta}
\end{equation}
\end{lemma}

Thus, by Lemma \ref{le:G_c}, we have
\begin{equation}
\begin{aligned}
    \left(\mathbf{F}(\mathbf{W}(t))-\mathbf{y}\right)^{\mathbf{T}}I_2
    &= -\eta 
    \left(\mathbf{F}(\mathbf{W}(t))-\mathbf{y}\right)^{\mathbf{T}}\tilde{\mathbf{G}}(t)(\mathbf{F}(\mathbf{W}(t))-\mathbf{y})
    \\
    & \leq
    \left(-\frac{\eta \lambda_0}{2}+\frac{\eta q n^{2} R}{\sqrt{2 \pi} \delta}\right)
    ||\mathbf{F}(\mathbf{W}(t))-\mathbf{y}||^{2}
\end{aligned}
\end{equation}
With the upper bound of the cardinal of $J_{true}(R)$(Lemma \ref{le:J}), we also have
\begin{equation}
\begin{aligned}
    \left(\mathbf{F}(\mathbf{W}(t))-\mathbf{y}\right)^{\mathbf{T}}I_1
    & \leq
    ||\mathbf{F}(\mathbf{W}(t))-\mathbf{y}||||I_1||
    \\
    & \leq
    ||\mathbf{F}(\mathbf{W}(t))-\mathbf{y}||\frac{\eta q \sqrt{n} |J_{true}(R)|}{M D}||\mathbf{F}(\mathbf{W}(t))-\mathbf{y}||
    \\
    & \leq
    \frac{q\eta n^{\frac{3}{2}} R}{\sqrt{2 \pi} \delta}
    ||\mathbf{F}(\mathbf{W}(t))-\mathbf{y}||^{2}
\end{aligned}
\end{equation}
Based on the estimation of upper bound of three terms, we have
\begin{equation}
\begin{aligned}
     ||\mathbf{F}(\mathbf{W}(t+1))-\mathbf{y}||^{2} 
    &= ||\mathbf{F}(\mathbf{W}(t+1))-\mathbf{F}(\mathbf{W}(t))||^{2}
    \\
    &\quad +2\left(\mathbf{F}(\mathbf{W}(t))-\mathbf{y}\right)^{\mathbf{T}}\left(\mathbf{F}(\mathbf{W}(t+1))-\mathbf{F}(\mathbf{W}(t))\right)
    +||\mathbf{F}(\mathbf{W}(t))-\mathbf{y}||^{2}
    \\
    &=
    ||\mathbf{F}(\mathbf{W}(t))-\mathbf{y}||^{2} +
    ||\mathbf{F}(\mathbf{W}(t+1))-\mathbf{F}(\mathbf{W}(t))||^{2}
    \\
    &\quad +2\left(\mathbf{F}(\mathbf{W}(t))-\mathbf{y}\right)^{\mathbf{T}}I_1 
    +2\left(\mathbf{F}(\mathbf{W}(t))-\mathbf{y}\right)^{\mathbf{T}}I_2
    \\
    & \leq
    \left(1+q^{2} {\eta}^{2} n^{2}-\eta\lambda_0 +\frac{2\eta q n^{2} R}{\sqrt{2 \pi} \delta}
    +\frac{2q\eta n^{\frac{3}{2}} R}{\sqrt{2 \pi} \delta} \right)
    ||\mathbf{F}(\mathbf{W}(t))-\mathbf{y}||^{2}
    \\
    & \leq
    \left(1-\frac{\eta\lambda_0}{2} \right)
    ||\mathbf{F}(\mathbf{W}(t))-\mathbf{y}||^{2}
\end{aligned}
\end{equation}
Finally, we only need to select the value of $R$. To make all the lemmas are true, we can set $R$ to $\frac{\sqrt{2 \pi} \delta \lambda_0}{16 q n^{2}}$. Now, by combining the inductive hypothesis, we prove Lemma \ref{le:main}.

Next, we prove Lemma \ref{le:W_fin} by the expansion of Lemma \ref{le: w_r(t)}. In fact, Lemma \ref{le:main} is true for any $t$, so Lemma \ref{le: w_r(t)} is true for any $T$. We use $\mathbf{W}_{fin,i} (1 \leq i \leq m)$ to denote the finetuned convolution filters and have $\mathbf{W}_{fin} = (\mathbf{W}_{fin,i})_{1 \leq i \leq m}$. Notice some filters have been pruned, by Lemma \ref{le: w_r(t)}, then we have
\begin{equation}
\begin{aligned}
    ||\mathbf{W}_{fin}-\mathbf{W}(0)||^{2}
    &=
    \sum_{r=1}^{M}||\mathbf{W}_{fin,r}-\mathbf{W}_r(0)||^{2}
    \\
    & \leq
    \sum_{r=1}^{M}\left(\frac{4\sqrt{q n}}{\sqrt{M} \lambda_0}
    ||\mathbf{F}(\mathbf{W}(0))-\mathbf{y}||\right)^{2}
    \\
    &=
    \frac{16q n}{\lambda_{0}^{2}}||\mathbf{F}(\mathbf{W}(0))-\mathbf{y}||^{2}
\end{aligned}
\end{equation}
By the concentration of random initialization, with high probability, $||\mathbf{F}(\mathbf{W}(0))-\mathbf{y}||$ is $O(\sqrt{n})$. So there exists an constant $C$ such that $||\mathbf{F}(\mathbf{W}(0))-\mathbf{y}||^{2} \leq C n$, which means we prove Lemma \ref{le:W_fin} .

\end{document}